\documentclass[11pt]{article}
\usepackage[final]{acl}

\usepackage{times}
\usepackage{latexsym}
\usepackage{bbding}
\usepackage{tcolorbox}
\tcbuselibrary{breakable}
\usepackage{colortbl}
\usepackage{xcolor}
\usepackage{subfig}
\usepackage{graphicx}
\usepackage{tcolorbox}
\tcbuselibrary{breakable}
\usepackage{colortbl,multirow}
\usepackage{bm}
\usepackage{makecell}
\usepackage{enumitem}
\usepackage{array}
\usepackage{caption}
\usepackage{adjustbox}
\usepackage{pifont}
\usepackage{hyperref}
\usepackage{amsmath}
\usepackage{booktabs}
\usepackage{amsfonts}
\usepackage{amssymb}
\usepackage{balance}
\usepackage{listings}

\definecolor{json_key}{rgb}{0, 0, 0.6}
\definecolor{json_string}{rgb}{0.1, 0.5, 0.1}
\definecolor{json_number}{rgb}{0.7, 0.4, 0}
\definecolor{code_background}{rgb}{0.98, 0.98, 0.98}
\definecolor{code_frame}{rgb}{0.85, 0.85, 0.85}

\lstdefinelanguage{json}{
    basicstyle=\ttfamily\small,
    numbers=left,
    numberstyle=\tiny\color{gray},
    stepnumber=1,
    numbersep=10pt,
    showstringspaces=false,
    breaklines=true,
    frame=single,
    backgroundcolor=\color{code_background},
    rulecolor=\color{code_frame},
    captionpos=b,
    stringstyle=\color{json_string},
    morestring=[s]{"}{"},
    literate=
     *{0}{{{\color{json_number}0}}}{1}
      {1}{{{\color{json_number}1}}}{1}
      {2}{{{\color{json_number}2}}}{1}
      {3}{{{\color{json_number}3}}}{1}
      {4}{{{\color{json_number}4}}}{1}
      {5}{{{\color{json_number}5}}}{1}
      {6}{{{\color{json_number}6}}}{1}
      {7}{{{\color{json_number}7}}}{1}
      {8}{{{\color{json_number}8}}}{1}
      {9}{{{\color{json_number}9}}}{1}
      {:}{{{\color{black}:}}}{1}
      {,}{{{\color{black},}}}{1}
      {\{}{{{\color{black}\{}}}{1}
      {\}}{{{\color{black}\}}}}{1}
      {[}{{{\color{black}[}}}{1}
      {]}{{{\color{black}]}}}{1},
}

\definecolor{limegreen}{rgb}{0.2, 0.8, 0.2}
\definecolor{lightblue}{rgb}{0.85,0.92,1}
\definecolor{lightyellow}{rgb}{1, 0.95, 0.8}
\definecolor{transgray}{gray}{0.92}

\newcommand{\dataname}{\textsc{RATs40K}}
\newcommand{\modelname}{\textsc{Time-RA}}

\usepackage[T1]{fontenc}
\usepackage[utf8]{inputenc}
\usepackage{microtype}
\usepackage{inconsolata}
\usepackage{graphicx}

\title{\modelname: Towards Time Series Reasoning for Anomaly Diagnosis \\ with LLM Feedback}

\author{
 \textbf{Yiyuan Yang\textsuperscript{1}\footnotemark[1]},
 \textbf{Zichuan Liu\textsuperscript{2}\footnotemark[1]},
 \textbf{Lei Song\textsuperscript{3}},
 \textbf{Kai Ying\textsuperscript{4}},
 \textbf{Zhiguang Wang\textsuperscript{5}}, 
\\
 \textbf{Tom Bamford\textsuperscript{6}},
 \textbf{Svitlana Vyetrenko\textsuperscript{6,7}},
 \textbf{Jiang Bian\textsuperscript{3}},
 \textbf{Qingsong Wen\textsuperscript{8}\footnotemark[2]}
\\
 \textsuperscript{1}University of Oxford,
 \textsuperscript{2}Nanjing University,
 \textsuperscript{3}MSRA,
 \textsuperscript{4}SJTU,
 \textsuperscript{5}Abel AI,
\\
 \textsuperscript{6}Outsampler,
 \textsuperscript{7}University of Strasbourg,
 \textsuperscript{8}Squirrel Ai Learning
\\
 \small{
   \textbf{Email:} \href{mailto:email@domain}{yiyuan.yang@cs.ox.ac.uk}, \href{mailto:email@domain}{zichuanliu@smail.nju.edu.cn}
 }
}

\begin{document}
\maketitle
\renewcommand{\thefootnote}{\fnsymbol{footnote}}
\footnotetext[1]{Equal Contribution.}
\footnotetext[2]{Corresponding Author.}
\renewcommand{\thefootnote}{\arabic{footnote}}
\begin{abstract}
Time series anomaly detection (TSAD) has traditionally focused on binary classification and often lacks the fine-grained categorization and explanatory reasoning required for transparent decision-making. To address these limitations, we propose Time-series Reasoning for Anomaly (\modelname), a novel task that reformulates TSAD from a discriminative into a generative, reasoning-intensive paradigm. To facilitate this, we introduce \dataname, the first real-world large-scale multimodal benchmark with \textasciitilde40,000 samples across 10 domains, integrating raw time series, textual context, and visual plots with structured reasoning annotations. Extensive benchmarking shows that while supervised fine-tuning and visual representations boost diagnostic accuracy and reasoning consistency, performance varies across complex scenarios. Notably, fine-tuned models demonstrate strong ``plug-and-play'' transferability, outperforming traditional baselines on unseen real-world datasets. Our work establishes a foundation for interpretable, multimodal time series analysis. All code\footnote{\url{https://github.com/yyysjz1997/Time-RA}} and the \dataname ~dataset\footnote{\url{https://huggingface.co/datasets/Time-RA/RATs40K}} are fully open-sourced to facilitate future research.
\end{abstract}

\section{Introduction}
Time series anomaly detection (TSAD) is critical across diverse domains, including finance, healthcare, AIOps, and industrial systems, where timely identification of anomalies prevents severe operational disruptions and economic losses~\cite{wang2024exploring,cook2019anomaly,ren2019time,khan2024anomaly,liu2024timex}. With the rapid development of artificial intelligence, especially deep learning (DL) and large language models (LLMs), significant advancements have been achieved in modeling complex temporal patterns and anomaly detection tasks~\cite{jin2024position}. However, the reasoning and detailed categorization of anomalies in time series remain underexplored, limiting our ability to reason diagnoses and hindering deeper understanding and informed decision-making based on identifying underlying causes beyond mere detection. In addition, existing TSAD benchmarks often lack explanatory reasoning and fine-grained anomaly categories for comprehensive diagnosis, creating a bottleneck for further advancement of TSAD~\cite{zhou2024llmsunderstandtimeseries}.

\begin{figure}[!t]
    \centering
    \includegraphics[width=1\linewidth]{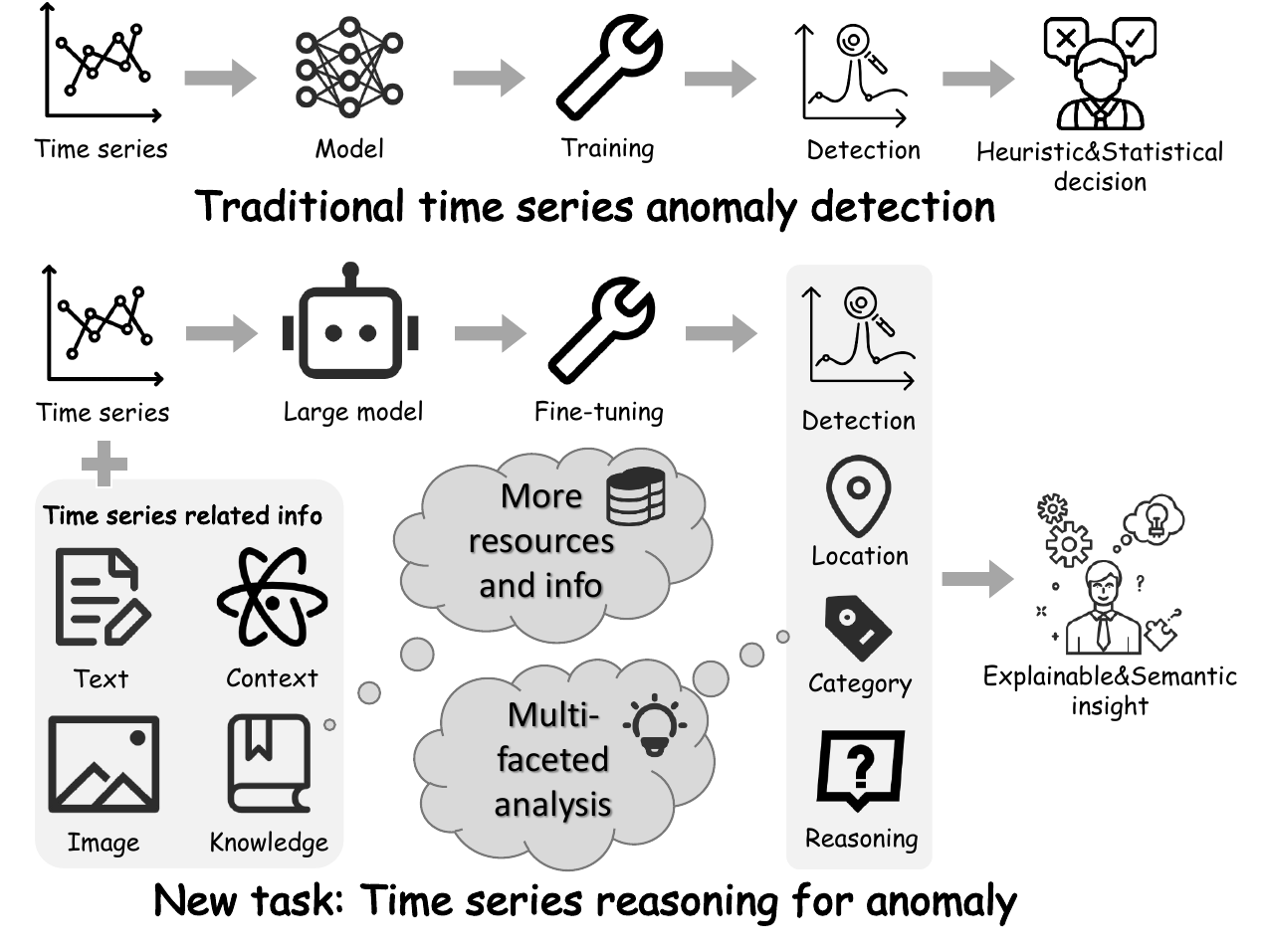}
    \caption{Comparison between traditional TSAD task and the proposed time series reasoning for anomaly task.}
    \label{fig:TSAD-task} 
\end{figure}

As shown in Figure~\ref{fig:TSAD-task}, the traditional TSAD task~\cite{yang2023dcdetector} focuses mainly on binary anomaly classification (anomalous vs. normal). It fails to provide the specific anomaly categories and diagnostic reasoning necessary for root-cause analysis. Also, existing researches on the TSAD task mainly focus on anomaly detection, neglecting the critical task of uncovering the underlying reasons behind these anomalies. Understanding these root causes is essential for future comprehensive decision-making processes, as it provides actionable insights and fosters interpretability for stakeholders~\cite{chow2024towards,liuexplaining}. The absence of in-depth causal analysis substantially limits practical utility, particularly in scenarios where preventive or corrective actions depend on anomaly origins, highlighting the need for a shift toward more comprehensive anomaly diagnosis.

Secondly, the scarcity of real-world datasets integrating multiple modalities (numeric time series data, textual explanations, and visual representations) hinders the advancement of multimodal TSAD and reasoning methods. Existing multimodal datasets are typically synthetic or limited to narrow contexts, inadequately capturing real-world complexity and variability~\cite{liu2025can,kong2025position}. Moreover, the reasoning capabilities of recent multimodal LLMs (MLLMs) remain underexplored due to a lack of high-quality annotated reasoning data. This gap limits the potential for explainable and interpretable anomaly detection outcomes, highlighting the pressing need for comprehensive, publicly available multimodal datasets and standardized benchmarks~\cite{chen2025mtbench}.

To address these limitations, we define a brand new task, \textbf{Time} series \textbf{R}easoning for \textbf{A}nomaly (\modelname), as shown in Figure~\ref{fig:TSAD-task}. This task converts discriminative models into generative models, e.g., LLMs and MLLMs, and performs domain-specific learning with multimodal inputs. Moreover, \modelname~reformats these inputs as structured prompts and fine-tunes LLMs, guiding them through a human-like diagnostic process structured into Observation, Thought, and Action stages. The task needs to output multi-objective results, including not only binary detection, but specific classes of anomalies and their reasons. This pipeline ensures clarity, consistency, and interpretability of the explanations generated by models.

To support \modelname~task, we introduce \textbf{R}easoning for \textbf{A}nomaly in \textbf{T}ime \textbf{s}eries \textbf{40K} (\dataname), the first real-world comprehensive multimodal dataset. It uniquely integrates numeric time series, contextual text, and visual representations, covering 10 real-world scenarios. It includes 14 univariate and 6 multivariate anomaly types, each with formal definitions, illustrative examples, and real-world scenarios for clarity and reproducibility. In addition, the initial annotations generated by a pool of the strongest LLMs currently available are subsequently refined through a rigorous AI-driven feedback process. This process results in annotations that are sufficiently accurate and suitable for supporting the \modelname~task.

Leveraging \dataname~dataset, we extensively fine-tune and evaluate multiple advanced LLMs, assessing their anomaly detection performance and reasoning capabilities across diverse anomaly categories and modalities. Through rigorous experimental evaluations, we derive meaningful insights into the strengths, limitations, and potential areas of improvement for current models, demonstrating the dataset's capability to support comprehensive benchmarking and foster substantial advancements in time series reasoning for anomaly task. In summary, our key contributions include: 

\begin{itemize}[left=0pt]
    \item \textbf{Task Reformulation (\modelname)}: We define \modelname, a novel task that elevates traditional binary detection into a generative diagnosis paradigm requiring joint detection, fine-grained categorization, and causal explanation.

    \item \textbf{Multimodal Dataset (\dataname)}: We construct \dataname, the first real-world multimodal reasoning dataset for \modelname ~with \textasciitilde40K samples across 10 domains, integrating raw time series, textual context, and visual plots with expert-aligned diagnostic labels. 

    \item \textbf{AI-Feedback Alignment Pipeline}: We develop a structured prompting and AI-assisted feedback pipeline that utilizes a diverse LLM pool and GPT-4 refinement to ensure high-quality, interpretable reasoning annotations.

    \item \textbf{Benchmarking and Generalization}: We provide extensive benchmarks of SOTA (M)LLMs, demonstrating that while fine-tuning and visual data boost reasoning quality, the task remains a complex frontier.
\end{itemize}

\begin{figure*}[h]
    \centering
    \includegraphics[width=1\linewidth]{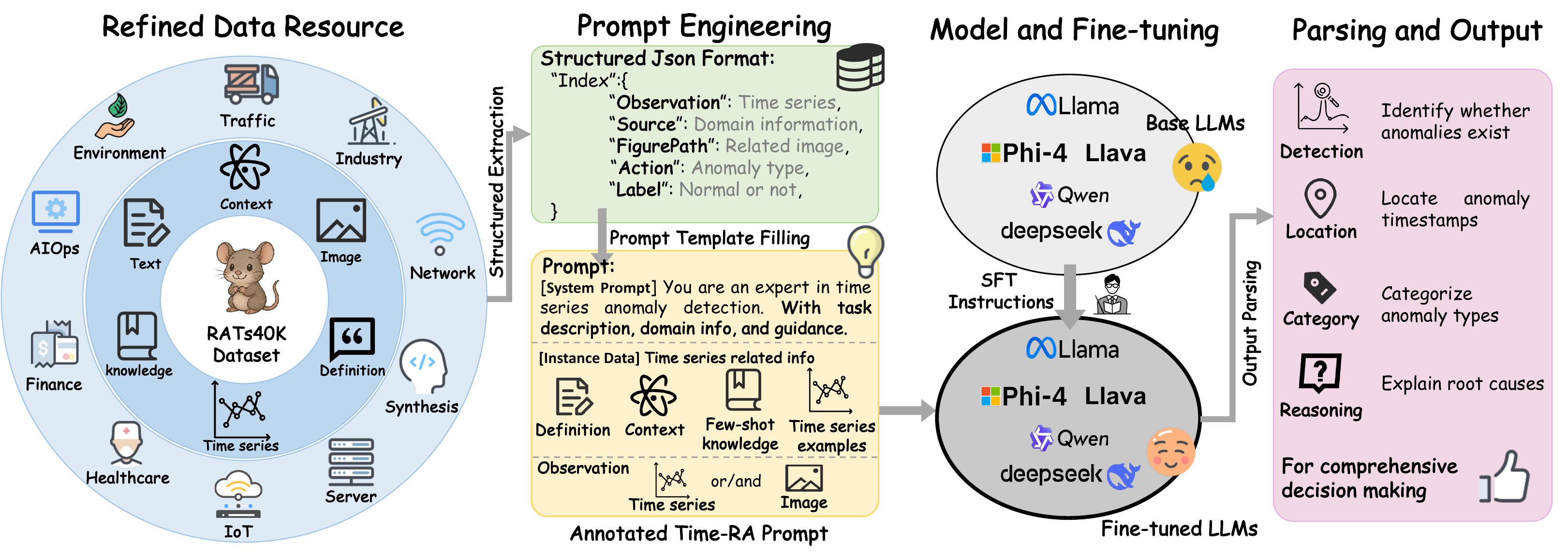}
    \caption{The end-to-end \modelname ~workflow highlights how multimodal inputs are structured and jointly optimized for detection, categorization, and reasoning.}
    \label{fig:timeRA}
\end{figure*}

\section{A New Lens for Anomaly Detection}
In this section, we introduce the Time series Reasoning for Anomaly~(\modelname) task, with an overview of its workflow presented in Figure~\ref{fig:timeRA}. We begin by curating a large-scale and diversified dataset, \dataname, which includes ground-truth of category and reason, detailed in the Section~\ref{dataisme}. This universal anomaly detection dataset is specifically formatted using our prompt engineering approach with a comprehensive anomaly definition, preparing it for model fine-tuning. Subsequently, the fine-tuned language model acquires enhanced expertise in time series analysis, enabling it to perform downstream anomaly detection tasks such as anomaly reasoning. Overall, \modelname~establishes a multi-tasking workflow for time series analysis, extending beyond the conventional binary anomaly detection problem~\cite{blazquez2021review} to address complex scenarios.

\subsection{Task Definition}
The \modelname~task extends traditional time series analysis by integrating multimodal reasoning and causal analysis. Formally, given multimodal input $\{T, D, V\}$, where $T$ is univariate/multivariate time series data, $D$ represents contextual textual metadata (e.g., domain descriptions), and $V$ is a visual representation. This task is multi-objective and requires a (vision) language model $\pi$ to perform:
\begin{itemize}[left=0pt]
\item \textbf{Anomaly Detection}: Identify whether the time series $T$ contains anomalous segments, i.e, determine an anomaly detection label $y_l = \pi_{\text{detect}}(\cdot|T, D, V)\in  \{0, 1\}$,
\item \textbf{Fine-grained Classification}: For anomalous segments, the model should also identify the specific class of anomaly so that we can invoke it as a model action. An action $a = \pi_{\text{classify}}(\cdot|T, D, V) \in \mathcal{C}_{\text{uni}} \cup \mathcal{C}_{\text{multi}}$, where $\mathcal{C}_{\text{uni}}$ and $\mathcal{C}_{\text{multi}}$ denote our comprehensive anomaly taxonomies in the Section \ref{anomalyca}.
\item \textbf{Model Thoughts}: The model also needs to generate human-understandable explanations, which usually contain the location of the anomaly and the model's thoughts for justifying both the anomaly presence and category assignment. Mathematically, a model's thought $r = \pi_{\text{reason}}(\cdot|T, D, V)$.
\end{itemize}
Based on the task definition, we first introduce expert-defined anomaly classifications and corresponding interpretations that provide a detailed reference for prompt engineering of anomaly data.

\subsection{Anomaly Category Definition}\label{anomalyca}
Traditional anomaly detection datasets mostly treat detection as a binary task (normal vs. anomalous) without distinguishing \textit{anomaly types}. In fact, identifying the category of an anomaly in practice is crucial for root cause analysis and effective decision-making. We therefore define a fine-grained taxonomy of anomaly classes and label each anomalous segment with its specific category. After surveying the literature and consolidating existing taxonomies~\cite{chandola2009anomaly,schmidl2022anomaly,blazquez2021review,liu2025detecting}, we select \textit{14 univariate anomaly types} and \textit{6 multivariate anomaly types} as our categorical labels. The univariate anomaly classes focus on time-localized deviations, such as point outliers, trend drifts, and nonlinear pattern anomalies. The multivariate anomaly classes consider both temporal and cross-series aberrations, for example, a trend deviation anomaly where one variable’s trend diverges significantly from others, or a joint-context anomaly where an otherwise acceptable pattern becomes anomalous when multiple variables are considered together. Each data segment in our dataset is thus annotated with both a binary label (normal or anomalous) and a specific anomaly category (or “normal” category for non-anomalous cases). In the Appendix, we provide detailed definitions for each anomaly type, along with example time series and real-world scenario descriptions to illustrate these categories in Tables~\ref{Table:Uni-clf-des} and~\ref{Table:Mul-clf-des}.

\subsection{Fine-tuning and Evaluation}
\textbf{Empowering LLMs}. Traditional anomaly detection methods, typically rooted in machine learning or deep learning, often face challenges in accurately classifying anomalies due to the scarcity of trustworthy labels and in providing meaningful causal analysis.  These limitations hinder their practical application in complex, real-world scenarios. More recently, the advent of generative pre-trained models such as LLMs has opened new avenues for time series anomaly detection~\cite{zhou2024llmsunderstandtimeseries,xu2025can}. Yet many of these initial explorations primarily focus on demonstrating LLMs' inherent capabilities without fully leveraging real-world, domain-specific data for enhanced learning. In \modelname~task, we address these limitations by fine-tuning LLMs with our unique \dataname~dataset. This approach empowers the LLMs with the ability to analyze time series anomalies more effectively and to generalize their understanding across diverse domains.

Mathematically, given a (vision) language model $\pi$ and after formatting an input $x=\{T, D, V\}$, we perform supervised fine-tuning as follows:
$$\max_{\theta} \mathbb{E}_{(x, y) \sim \mathcal{D}_{\dataname}} [\log P_{\theta}(y|x)].
$$
Here, $\theta$ represents the parameters of the language model $\pi$, $(x,y)$ are input-output pairs from the \dataname, and $y=\{y_l, a, r\}$. This fine-tuning process allows the LLM to learn the intricate patterns and contextual information present in our real-world dataset, thereby improving its ability to identify and interpret anomalies.

To thoroughly assess the performance of our fine-tuned $\text{\modelname}$ model, we adopt a multifaceted evaluation approach, considering three distinct aspects of anomaly detection: binary anomaly classification, multi-class anomaly type identification, and textual reasoning and explanation generation.

\textbf{Evaluation}. To thoroughly assess the performance of our fine-tuned models, we consider three types of evaluation separately: numerical binary label $y_l$ for determining whether a point is anomalous, numerical multi-class action $a$ for identifying the type of anomaly, and textual reasoning $r$ for explanation. For the first two quantitative evaluations, we treat them as classification problems and employ standard metrics such as precision (\textit{P}), recall (\textit{R}), and F1-score (\textit{F1}) for evaluation. For the evaluation of reasoning, we employ text similarity-based metrics, including \textit{Cosine} Similarity based on ELECTRA~\cite{clark2020electra}, \textit{TF-IDF} Similarity, \textit{Levenshtein} Similarity, \textit{Token} Sequence Similarity, and Reasoning Consistency Score (\textit{RCS}). For RCS, we employ \texttt{GPT-4} as an evaluator to score the logical integrity of the generated reasoning on a scale of 1 to 5, where a higher score indicates superior logical consistency and fewer hallucinations.

\begin{figure*}[!t]
    \centering
    \includegraphics[width=1\linewidth]{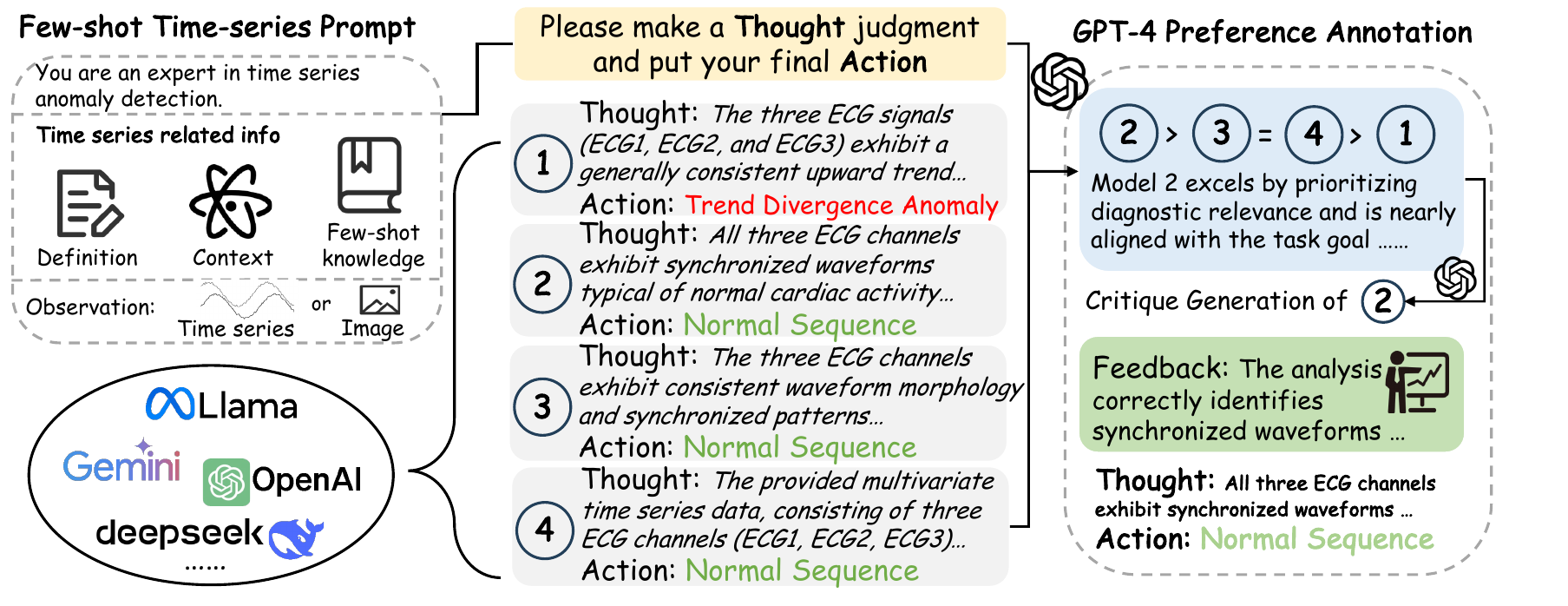}
    \caption{\dataname~dataset construction pipeline. 
    We ensure diversity by sampling anomalous thoughts and decision actions from a large model pool, then we query \texttt{GPT-4} with detailed definitions and prompts for preference selection and correction to generate high-quality, fine-grained annotations.}
    \label{fig:mainly}
\end{figure*}

\section{A New Dataset for \modelname ~Task}\label{dataisme}
To complete the \modelname~task, we establish a first anomaly reasoning dataset \dataname, including scalability and diversity of anomalies for model generalization. In this section, we integrate these objectives into all three phases: data collection, generating completions, and preference annotation inspired by the data engineering principles~\cite{cui24f,chiang2023vicuna}. As illustrated in Figure~\ref{fig:mainly} and compared with existing datasets in Table~\ref{comparison}, \dataname~emphasizes abnormal sample collection, enabling language models to capture heterogeneous anomaly types and their underlying causes. The details are provided in Appendix~\ref{details_dataset}.

\subsection{Dataset Collection}
We first collect multimodal time series anomaly data, incorporating numeric time series, descriptive text, and visualization images. 
The description of data sources is deferred to the Table~\ref{tab:datasets}.
Numerous time series segments of different lengths were extracted from these sources and labeled for anomalies.
We then attach a brief textual description to each sample, providing its application context and explaining the meaning of each feature in the multivariate series. Furthermore, each time series segment is rendered as a visual chart to facilitate comparative analysis of temporal patterns across variables. Finally, we obtain nearly 40K segments from various modalities across multiple fields, aiming to replicate how human analysts combine data sources for interpretation.

\subsection{Reason Completion Sampling}
To generate initial reasoning labels for our dataset, we design a structured prompting strategy that guides an LLM through a step-by-step anomaly analysis. The prompt adopts a role-based instruction, explicitly casting the model as an ``expert in time series anomaly detection''. It formalizes the task into three stages, \textbf{Observation}, \textbf{Thought}, and \textbf{Action}, mimicking the diagnostic reasoning process of human analysts. In the \{Observation\} stage, the prompt presents the time series data along with its domain knowledge. The model is then asked to articulate its \{Thought\}: a detailed reasoning process examining the time series behavior, relationships among variables, and any deviations from normal patterns. Finally, the model must output an \{Action\}, which in this context is the decided anomaly category from our predefined list. The prompt includes the full list of anomaly category options with brief natural-language definitions, also a few exemplary question-answer cases serving as in-context demonstrations. The resultant structured prompt for univariate/multivariate anomaly detection is formalized in Appendix~\ref{Annotation}.

Using the above prompt setup, we leverage LLMs to automatically annotate each time series segment with a reasoning explanation and an anomaly category. More specifically, we built a model pool that consists of $4$ powerful models in current arena leaderboards: \texttt{GPT-4o}~\cite{achiam2023gpt}, \texttt{Gemini-2.5}~\cite{team2023gemini}, \texttt{DeepSeek-R1}~\cite{guo2025deepseek}, and \texttt{Llama-3.3-70B-Instruct}~\cite{grattafiori2024llama}. Then, the models are given the Observation and domain context, and they produce a Thought $r$ and an Action $a$ of the time series segment, where statistics of responses are in the Appendix~\ref{llm-com-section}. This yielded a preliminary set of anomaly annotations with human-readable explanations. In the subsequent steps, we further refine these LLM-generated annotations through an iterative feedback process to ensure the final labels are of high quality.

\begin{table*}[!t]
\centering
\caption{Statistics of existing time series anomaly datasets. "-" denotes that the corresponding data does not exist or missing classification and reasoning labels.
 }
    \resizebox{\textwidth}{!}{
\begin{tabular}{@{}lcccccccccc@{}}
\toprule
\multicolumn{1}{l|}{\textbf{Dataset}}      & \textbf{\# Sample} & 
\textbf{Modalities} & \textbf{\# Domain} & \textbf{\begin{tabular}[c]{@{}c@{}}Is Real-\\World?\end{tabular}}  & \textbf{\begin{tabular}[c]{@{}c@{}}Time series\\ Dimension\end{tabular}} & \textbf{\begin{tabular}[c]{@{}c@{}}Anomaly\\ Ratio\end{tabular}} & \textbf{\begin{tabular}[c]{@{}c@{}}\# Anomaly\\ Category\end{tabular}} & \textbf{\begin{tabular}[c]{@{}c@{}}Thought\\ Length\end{tabular}} & \textbf{Annotation} \\ \midrule
\multicolumn{1}{l|}{\textbf{AnomLLM}~\cite{zhou2024llmsunderstandtimeseries}}& 3,200& Time+Text & - & \XSolidBrush & U & 64.5\% & 8 & -  & Synthetic  \\
\multicolumn{1}{l|}{\textbf{LLMAD}~\cite{liu2024large2}} & ~37,000 & Time+Text+Image  & 3 & \Checkmark  & U & 0.71\%- 2.35\% & 8 & - & 100 by Human  \\
\multicolumn{1}{l|}{\textbf{VisualTimeAnomaly}~\cite{xu2025can}} &12,400& Time+Text+Image  & - & \XSolidBrush & U \& M & - & 9 & -  & Synthetic \\
\midrule
\multicolumn{1}{l|}{\textbf{\dataname} (Ours)}         & 39,574  &  Time+Text+Image & 10 & \Checkmark  &  U \& M & 83.7\% & 14 + 6  & 101.378  & AI feedback\\ \bottomrule
\end{tabular}
}\label{comparison}
\end{table*}

\subsection{AI Feedback Annotation}
After generating 158,340 model completions from 36,311 univariate and 3,274 multivariate instructions, we use \texttt{GPT-4} to provide two kinds of feedback for each: quantitative rankings that pinpoint a model's placement in the model pool, and textual critiques that provide detailed suggestions for the thought of anomaly detection. We leverage \texttt{GPT-4} for feedback generation due to the limited scalability of human evaluation and the potential subjectivity introduced by human annotators. In total, this resulted in more than 150,000 feedback data points. To ensure the quality of the model responses in the pool, we identify preferences and conduct a critique, selecting the top-ranked results as the final dataset. Furthermore, to validate the reliability of the LLM-generated feedback, we conduct expert evaluations on a subset of the results in the Section~\ref{data_quality_sec}, which confirmed the alignment between \texttt{GPT-4} assessments and human judgments. We put the \texttt{GPT-4} annotation and critique instruction in Appendix~\ref{Annotation}.

\textbf{Preference Annotation}. To enhance the reliability of GPT-4's annotations and minimize subjectivity and randomness, we implement three key techniques~\cite{cui24f}: (i) Reference: For each type of time series anomaly, we provide expert classification and corresponding reasons in the instruction as the model's few-shot, which helps reduce randomness. (ii) Standardization: For each of these aspects, we gave \texttt{GPT-4} a clear Likert scale with scores from 1 to 5 as a reference. (iii) Reasoning: In addition to ranking the models, \texttt{GPT-4} also provides a reason to explain how to rate each model based on abnormal time series. Combining all techniques, we ultimately obtain scalar scores and reasons for each model response. Then, we also use \texttt{GPT-4} for preference ranking in the model pool and select the anomaly detection answer with the best score as a label for each sample. 

\textbf{Critique Generation}. To complement our scalar scores, we leverage \texttt{GPT-4} for textual critiques. Our goal is to enable the LLM to serve as a mentor, providing specific and detailed advice for the top-ranked model in each ranking to guide its improvement and placement in the final dataset, rather than simply providing direct preference rankings. These prompts are unique for each anomaly time series and are generated from the overall perspective of the model pool, thus serving as the optimal reasoning for anomalies.  After generating feedback through \texttt{GPT-4}, we merge the feedback results with the top-ranked results in the model pool, and place them in the final \dataname~dataset to improve the label quality of anomaly reason.

\begin{figure}[!t]
    \centering
    \includegraphics[width=1\linewidth]{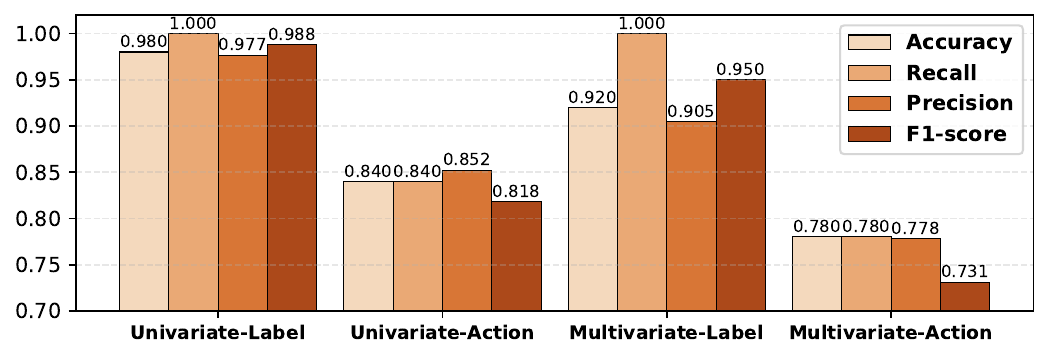}
    \caption{Comparison of LLM-generated labels with expert annotations.}
    \label{fig:data_quality-label} 
\end{figure}

\begin{figure}[!t]
    \centering
    \subfloat[Univariate quality for the textual reasoning.]{%
       \includegraphics[width=1\linewidth]{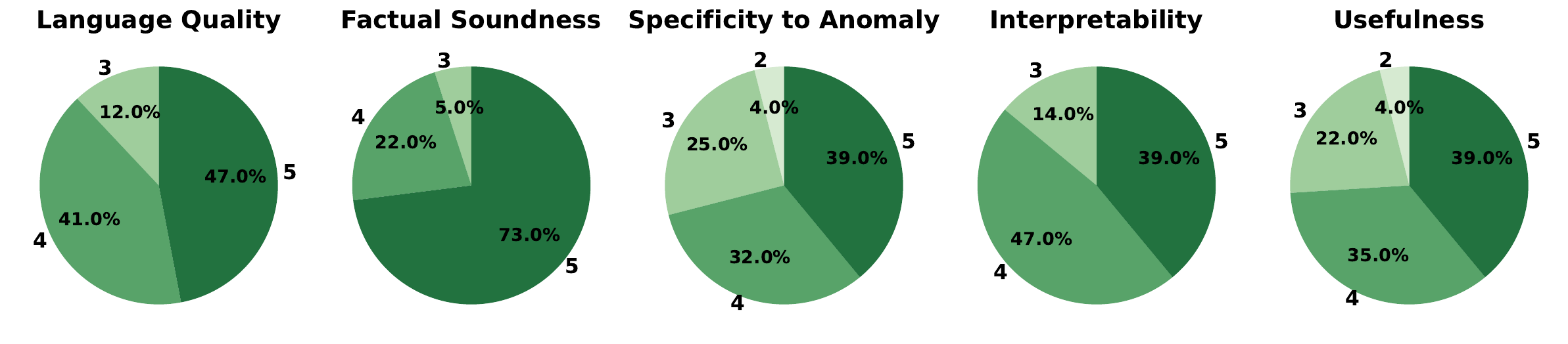}
       \label{fig:po-comparison-uni}
    } 
    \\ 
    \subfloat[Multivariate quality for the textual reasoning.]{%
      \includegraphics[width=1\linewidth]{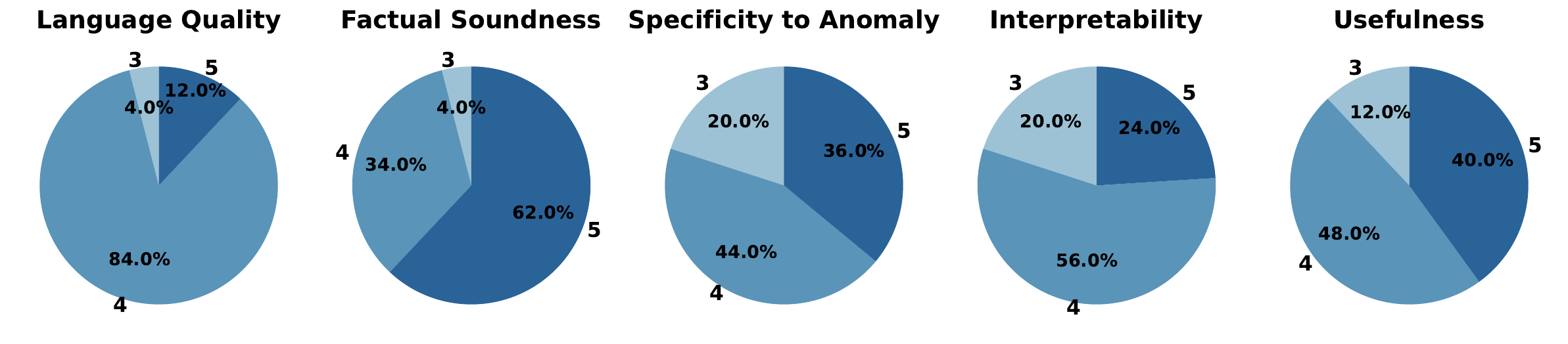}
      \label{fig:po-comparison-multi}
    }
    \caption{Textual reasoning quality based on Likert scale (Based on 100 uni- and 50 multivariate samples).}
    \label{fig:po-comparison}
\end{figure}

\section{Experiments}
In this section, we first conduct an in-depth quality analysis of \dataname. Then, we present the performances of LLMs/MLLMs on it and provide insights to guide future work on the \modelname~task. 

\subsection{Settings}
We evaluate some widely used open-source LLMs/MLLMs under both zero-shot and supervised fine-tuning (SFT) settings. The detailed list of models with their links can be found in Appendix~\ref{LLM_describe} and Table~\ref{llms}. For the SFT, each sample is formatted based on a fixed instruction template, and we use LoRA~\cite{hu2022lora} for parameter-efficient fine-tuning. In detail, full training configurations and hyperparameters are provided in our source code to ensure reproducibility. For fine-tuning, we adopt LoRA with rank $r = 8$, scaling factor $\alpha = 32$, and dropout rate 0.05, applied to the $qkv_{\text{proj}}$ and $o_{\text{proj}}$ modules. Models are trained in \texttt{bfloat16} precision using the AdamW optimizer with a learning rate of $5 \times 10^{-6}$. To evaluate the model outputs, we design regular expressions to automatically extract the predicted Thought and Action, which will then be compared against the ground truth. Evaluation metrics follow the definition of the \modelname~task, with the best results in \textbf{bold} and second-best \underline{underline}.

\begin{table*}[ht]
\centering
\caption{Performance on time-series-only anomalies based on LLM. Pure color rows are the machine learning or LLM result, while \colorbox {gray!15}{gray rows} represent the performance after fine-tuning. F1 means weighted-F1 score.}
\renewcommand{\arraystretch}{1}
\resizebox{\textwidth}{!}{
\rowcolors{2}{gray!15}{white}
\begin{tabular}{c|c|cccc|cccc|ccccc}
\toprule
& & \multicolumn{4}{c|}{\textbf{Label Matching}} & \multicolumn{4}{c|}{\textbf{ActionID Matching}} & \multicolumn{5}{c}{\textbf{Thought Matching}} \\ \cline{3-6} \cline{7-10} \cline{11-15}
\cellcolor{white}    &  \cellcolor{white} Model &  \cellcolor{white}P &  \cellcolor{white}R &  \cellcolor{white}F1 &  \cellcolor{white}Rank &  \cellcolor{white}P &  \cellcolor{white}R &  \cellcolor{white}F1 &  \cellcolor{white}Rank &   \cellcolor{white}Cosine& \cellcolor{white}TFIDF& \cellcolor{white}Lev. & \cellcolor{white}Token& 
\cellcolor{white}RCS \\
\midrule
 \cellcolor{white} & DeepSeek-7B & 0.8175 & 0.3295 & 0.4697 & 10th & 0.1190 & 0.1407 & 0.0703 & 10th & 0.8780 & 0.2026 & 0.1731 & 0.1001 & 2.1746 \\
 \cellcolor{white} & DeepSeek-7B & 0.8598 & 0.2729 & 0.4143 & 12th & 0.1532 & 0.1613 & 0.0771 & 9th & 0.8971 & 0.2298 & 0.1773 & 0.1039 & 2.3524 \\
 \cellcolor{white} & Llama-3-8B  & 0.9098 & 0.3148 & 0.4678 & 11th & 0.1767 & 0.1754 & 0.0897 & 8th & 0.9205 & 0.3154 & 0.2569 & 0.1533 & 2.2330 \\
 \cellcolor{white} & Llama-3-8B  & 0.8576 & 0.7605 & 0.8061 & 8th & 0.3887 & 0.1650 & 0.1511 & \textbf{1st} & 0.9242 & 0.3065 & 0.2484 & 0.1511 & 2.6780 \\
\cellcolor{white} Univariate & Llama-3.2-3B & 0.8510 & 0.3946 & 0.5391 & 9th & 0.2353 & 0.1518 & 0.0916 & 7th & 0.9127 & 0.2752 & 0.2295 & 0.1467 & 2.1363 \\
\cellcolor{white} Time Series & Llama-3.2-3B & 0.8458 & 0.8447 & 0.8453 & 5th & 0.1846 & 0.1038 & 0.0986 & 6th & 0.9264 & 0.3175 & 0.2524 & 0.1613 & 2.3111 \\
 \cellcolor{white} Anomaly & Phi-4-mini  & 0.8631 & 0.8062 & 0.8337 & 7th & 0.2141 & 0.1428 & 0.1309 & 3rd & 0.9166 & 0.3208 & 0.2187 & 0.1525 & 3.2840 \\
 \cellcolor{white} Detection & Phi-4-mini & 0.8582 & 0.8158 & 0.8364 & 6th & 0.2035 & 0.1482 & 0.1397 & \underline{2nd} & 0.9168 & 0.3154 & 0.2163 & 0.1518 & 3.2633 \\
 \cellcolor{white} & Qwen2.5-3B & 0.8296 & 0.9834 & 0.9000 & \textbf{1st} & 0.2016 & 0.0605 & 0.0301 & 12th & 0.9108 & 0.2841 & 0.2129 & 0.1306 & 2.0446 \\
 \cellcolor{white} & Qwen2.5-3B & 0.8299 & 0.9768 & 0.8974 & \underline{2nd} & 0.1592 & 0.0623 & 0.0371 & 11th & 0.9118 & 0.2849 & 0.2131 & 0.1316 & 2.3682 \\
 \cellcolor{white} & Qwen2.5-7B & 0.8465 & 0.9352 & 0.8886 & 3rd & 0.2436 & 0.1112 & 0.1064 & 5th & 0.9253 & 0.3059 & 0.2435 & 0.1619 & 3.2422 \\
 \cellcolor{white} & Qwen2.5-7B & 0.8452 & 0.9295 & 0.8854 & 4th & 0.2532 & 0.1229 & 0.1100 & 4th & 0.9270 & 0.3160 & 0.2432 & 0.1624 & 3.2120 \\
\midrule
 \cellcolor{white} & DeepSeek-7B & 0.6980 & 0.4419 & 0.5411 & 8th & 0.0626 & 0.1224 & 0.0821 & 12th & 0.9344 & 0.3420 & 0.2305 & 0.1190 & 2.2534 \\
 \cellcolor{white} & DeepSeek-7B & 0.7112 & 0.4264 & 0.5331 & 9th & 0.0658 & 0.1344 & 0.0874 & 11th & 0.9344 & 0.3358 & 0.2267 & 0.1207 & 2.2811 \\
 \cellcolor{white} & Llama-3-8B & 0.7736 & 0.3067 & 0.4393 & 12th & 0.4491 & 0.2114 & 0.1303 & 9th & 0.9398 & 0.3869 & 0.2604 & 0.1481 & 2.1926 \\
 \cellcolor{white} & Llama-3-8B & 0.8176 & 0.3668 & 0.5064 & 11th & 0.5934 & 0.1803 & 0.1151 & 10th & 0.9455 & 0.3887 & 0.2625 & 0.1459 & 2.2610 \\
 \cellcolor{white} Multivariate & Llama-3.2-3B & 0.8345 & 0.4515 & 0.5860 & 7th & 0.5008 & 0.2842 & 0.2583 & 7th & 0.9580 & 0.4479 & 0.2805 & 0.1698 & 2.8624 \\
 \cellcolor{white}Time Series & Llama-3.2-3B & 0.8232 & 0.4963 & 0.6193 & 6th & 0.4838 & 0.3016 & 0.2893 & 6th & 0.9578 & 0.4481 & 0.2788 & 0.1682 & 2.8934 \\
 \cellcolor{white}  Anomaly & Phi-4-mini & 0.7676 & 0.8237 & 0.7947 & 3rd & 0.4834 & 0.4056 & 0.4339 & \underline{2nd} & 0.9385 & 0.3842 & 0.2433 & 0.1319 & 2.4657 \\
 \cellcolor{white}Detection & Phi-4-mini & 0.7622 & 0.8195 & 0.7899 & 4th & 0.4627 & 0.4230 & 0.4372 & \textbf{1st} & 0.9376 & 0.3776 & 0.2399 & 0.1313 & 2.5389 \\
 \cellcolor{white} & Qwen2.5-3B & 0.6339 & 0.4551 & 0.5299 & 10th & 0.2397 & 0.2180 & 0.2209 & 8th & 0.8822 & 0.2901 & 0.1808 & 0.0892 & 2.3759 \\
 \cellcolor{white} & Qwen2.5-3B & 0.6545 & 0.6039 & 0.6281 & 5th & 0.3563 & 0.2876 & 0.2933 & 5th & 0.8762 & 0.2893 & 0.1738 & 0.0937 & 2.6421 \\
 \cellcolor{white} & Qwen2.5-7B & 0.7567 & 0.9801 & 0.8541 & \underline{2nd} & 0.5256 & 0.4084 & 0.4127 & 4th & 0.9532 & 0.3792 & 0.2676 & 0.1499 & 2.7454 \\
 \cellcolor{white} & Qwen2.5-7B & 0.7557 & 0.9826 & 0.8544 & \textbf{1st} & 0.5675 & 0.4322 & 0.4300 & 3rd & 0.9554 & 0.3805 & 0.2689 & 0.1504 & 2.7033 \\
\bottomrule
\end{tabular}}
\label{llm-result-main}
\end{table*}

\begin{table*}[!t]
\centering
\caption{Performance on visualized time series anomalies based on MLLMs. Pure color rows are the model's observations solely through images, while \colorbox {gray!15}{gray rows} are observations that include both images and time series.
}
\renewcommand{\arraystretch}{0.95}
\resizebox{\textwidth}{!}{
    \rowcolors{2}{gray!15}{white} 
    \begin{tabular}{c|c|cccc|cccc|ccccc}
    \toprule
& & \multicolumn{4}{c|}{\textbf{Label Matching}} & \multicolumn{4}{c|}{\textbf{ActionID Matching}} & \multicolumn{5}{c}{\textbf{Thought Matching}} \\
\cline{3-6} \cline{7-10} \cline{11-15}
 \cellcolor{white}    &  \cellcolor{white} Model &  \cellcolor{white}P &  \cellcolor{white}R &  \cellcolor{white}F1&  \cellcolor{white}Rank &  \cellcolor{white}P &  \cellcolor{white}R &  \cellcolor{white}F1 &  \cellcolor{white}Rank &   \cellcolor{white}Cosine& \cellcolor{white}TFIDF& \cellcolor{white}Lev.& \cellcolor{white}Token& \cellcolor{white}RCS \\
    \midrule
 \cellcolor{white}&Llava-v1.5-7B&0.8272&0.9489&0.8839&\underline{2nd}&0.1722&0.0358&0.0231&8th&0.9274&0.3244&0.2299&0.1450&1.9572\\
 \cellcolor{white} &Llava-v1.5-7B&0.8261&0.9668&0.8909&\textbf{1st}&0.1933&0.0579&0.0580&7th&0.9195&0.2876&0.2197&0.1242&2.7730\\
 \cellcolor{white} Univariate &Llava-v1.5-13B&0.8268&0.7601&0.7921&7th&0.1717&0.1931&0.1693&\textbf{1st}&0.9083&0.2995&0.2386&0.1143&2.1874\\
 \cellcolor{white} Time Series &Llava-v1.5-13B&0.8216&0.9165&0.8665&4th&0.1781&0.1572&0.1128&\underline{2nd}&0.9355&0.3436&0.2725&0.1301&2.3602\\
 \cellcolor{white} Anomaly &Llama-3.2-11B-v &0.8322&0.8050&0.8183&6th&0.1866&0.1030&0.0863&4th&0.8921&0.2483&0.1586&0.0956&2.1240\\
 \cellcolor{white} Detection &Llama-3.2-11B-v &0.8216&0.9165&0.8665&4th&0.1725&0.0944&0.0856&5th&0.9155&0.2834&0.1954&0.1072&2.7210\\
 \cellcolor{white} &Qwen2.5-vl-7B&0.8281&0.3977&0.5373&8th&0.0413&0.1566&0.0651&6th&0.8998&0.2088&0.1698&0.1177&1.9824\\
 \cellcolor{white} &Qwen2.5-vl-7B&0.8426&0.8973&0.8691&3rd&0.1171&0.1128&0.0930&3rd&0.9011&0.2685&0.1873&0.1297&2.7376\\
    \midrule
 \cellcolor{white} &Llava-v1.5-7B&0.9174&0.2532&0.3968&8th&0.5970&0.2618&0.1569&4th&0.9130&0.3755&0.2642&0.1125&2.1634\\
 \cellcolor{white}  &Llava-v1.5-7B&0.7340&0.4340&0.5455&7th&0.4715&0.1238&0.1150&6th&0.9359&0.3959&0.2529&0.1266&2.2750\\
 \cellcolor{white} Multivariate &Llava-v1.5-13B&0.7287&0.6919&0.7098&5th&0.3716&0.4216&0.3944&\textbf{1st}&0.9451&0.3628&0.2525&0.1160&2.3619\\
 \cellcolor{white} Time Series &Llava-v1.5-13B&0.8058&0.8682&0.8358&\underline{2nd}&0.6308&0.2036&0.2880&3rd&0.9449&0.3869&0.2479&0.1267&2.3647\\
 \cellcolor{white} Anomaly &Llama-3.2-11B-v&0.8012&0.4164&0.5480&6th&0.2402&0.1986&0.1300&5th&0.9254&0.3088&0.2200&0.1108&2.1667\\
 \cellcolor{white} Detection &Llama-3.2-11B-v & 0.8571 & 0.7500 & 0.8000 &3rd&0.4091&0.2727&0.2922&\underline{2nd}&0.9350& 0.3429& 0.2219&0.1129&3.3636\\
 \cellcolor{white} &Qwen2.5-vl-7B&0.7381&1.0000&0.8493&\textbf{1st}&0.0039&0.0568&0.0073&8th&0.9307&0.3079&0.2348&0.1296&1.5842\\
 \cellcolor{white} &Qwen2.5-vl-7B&0.7627&0.7816&0.7721&4th&0.3728&0.1245&0.1077&7th&0.9526&0.3994&0.2651&0.1419&2.5886\\
    \bottomrule
\end{tabular}
}
\label{tab_pass1-acc}
\end{table*}

\subsection{Reliability of the \dataname~Dataset} \label{data_quality_sec}
Following LLMAD~\cite{liu2024large2}, we assess label quality by comparing 7 human experts' average annotations (both binary and action classification) with LLM-based generated ranking results, as shown in Figure~\ref{fig:data_quality-label}. For both univariate and multivariate samples, LLM-based labels show high consistency with expert annotations. Binary classification results demonstrate strong agreement with high accuracy and F1-scores, while action classification gets slightly lower scores, reflecting the increased complexity and uncertainty of fine-grained anomaly categorization. As for the quality of the textual reasoning, we evaluate five key dimensions inspired by human explanation quality criteria based on a well-designed Likert scale. A more detailed description can be found in Appendix~\ref{data_quality}. For the univariate results from Figure~\ref{fig:po-comparison-uni}, the average scores across these dimensions range from 4.04 to 4.58 (out of 5), indicating consistent clarity, factual alignment, and actionable insights. The multivariate subset shows similarly high quality, with average scores from 4.08 to 4.28, as shown in Figure~\ref{fig:po-comparison-multi}. These results further demonstrate the high quality of our annotated data.

\subsection{Results and Discussion}
For the results and evaluation, we focus on four research questions for the \modelname~task as follows:

\textbf{RQ 1: Can LLMs adapt to the number of variates?} Based on the prompt design and input architecture discussed previously, the results show in Tables~\ref{llm-result-main}, \ref{tab_pass1-acc}, \ref{other-data-01clf}, and \ref{other-alg-01clf-univar} indicate that the LLMs can effectively adapt to the varying number of variates. Specifically, models such as \texttt{Qwen2.5} show comparable or better F1-scores on multivariate tasks (\textasciitilde{}0.85) compared to univariate tasks (\textasciitilde{}0.9). Besides, LLM can adapt to different numbers of variables (from 2 to 9) in Tables~\ref{llm-result-main} and~\ref{tab_pass1-acc} and even the 17 and 19 dimensions in Table~\ref{other-data-01clf}. Overall, these quantitative insights suggest that an appropriate prompt engineering design may help LLMs better leverage the additional complexity inherent in multivariate time series. It indicates promising adaptability across varying numbers of variables.

\textbf{RQ 2: Can SFT enhance LLMs' performance on \modelname ~task using \dataname?} Analysis of Table~\ref{llm-result-main} indicates that SFT generally improves performance, with most models achieving higher F1-scores and better semantic alignment in Thought Matching after tuning. However, these gains are not universal. In some complex multivariate scenarios, performance remains stagnant or even slightly regresses. This inconsistency underscores the inherent difficulty of the \modelname ~task, particularly when dealing with the noise and complexity of real-world data. These results suggest that while SFT is a beneficial baseline, it may not be sufficient to fully resolve the intricacies of the task. The gap between LLMs and specialized supervised models highlights the need for more sophisticated and specialized adaptation methods to achieve robust reliability in practical applications.

\textbf{RQ 3: Can visual representation enhance the \modelname ~task?} Table~\ref{tab_pass1-acc} demonstrates that visual representation is a powerful catalyst for the \modelname ~task, particularly in enhancing the depth and coherence of model reasoning. The obvious gains are observed in Thought Matching metrics, where visual plots consistently help models align their internal logic with task-specific actions, achieving higher semantic scores. While the impact on classification varies by model, the integration of visual and raw data allows models like \texttt{Llama-3.2-11B-v} to reach peak performance. These results suggest that visualization effectively bridges the gap between raw data and high-level reasoning. Although the complexity of multimodal fusion remains a challenge, the clear improvement in reasoning quality confirms that visual representation is instrumental in guiding MLLMs toward more accurate and interpretable anomaly diagnostics.

\begin{table}[]
\centering
\caption{Performances of the recent and ~\dataname ~fine-tuned LLM based on other real-world datasets (in-domain and out of domain) for the anomaly detection task. GHL has 19 channels. CATSv2 has 17 channels.} 
\renewcommand{\arraystretch}{1}
\resizebox{1\linewidth}{!}{
\begin{tabular}{c|c|ccc|c|ccc}
\toprule
\multirow{2}{*}{Dataset} & \multirow{2}{*}{Model} & \multicolumn{3}{c|}{\textbf{Univariate TS}} & \multirow{2}{*}{Dataset} & \multicolumn{3}{c}{\textbf{Multivariate TS}} \\ \cline{3-5} \cline{7-9} 
 &  & P & R & F1 &  & P & R & F1 \\ \midrule
 & KNN & 0.9167 & 0.1571 & 0.2683 &  & 0.5625 & 0.1023 & 0.1731 \\
 & LOF & 0.8333 & 0.0714 & 0.1316 &  & 0.4000 & 0.0682 & 0.1165 \\
 & AE1SVM & 0.3333 & 0.0571 & 0.0976 &  & 0.5000 & 0.0795 & 0.1373 \\
 & XGBoost & 0.9429 & 0.9429 & 0.9429 &  & 0.9053 & 0.9773 & 0.9399 \\
NEK & LSTM & 0.6796 & 1.0000 & 0.8092 & GHL & 0.5252 & 0.8295 & 0.6432 \\
(In) & TimesNet & 0.7130 & 0.8034 & 0.7556 & (In) & 0.4843 & 0.7998 & 0.6034\\
 & AT & 0.7531 & 0.8998 & 0.8200 &  & 0.5000 & 0.8129 & 0.6193\\
 & Timer & 0.7370 & 0.7175 & 0.7272 &  & 0.5276 & 0.5998 & 0.5615\\
 & Chronos & 0.7579 & 0.7680 & 0.7630 &  & 0.5778 & 0.6921 & 0.6299\\
 \rowcolor{gray!15} \cellcolor{white} & Qwen2.5-7B & 0.5397 & 0.9855 & 0.6974 & \cellcolor{white} & 0.8923 & 0.6591 & 0.7582  \\ \midrule
 & KNN & 1.0000 & 0.1569 & 0.2712 &  & 0.7333 & 0.1000 & 0.1760 \\
 & LOF & 0.9583 & 0.2255 & 0.3651 &  & 0.5600 & 0.1273 & 0.2074 \\
 & AE1SVM & 0.9444 & 0.1667 & 0.2833 &  & 0.6364 & 0.1273 & 0.2121 \\
 & XGBoost & 0.9806 & 0.9902 & 0.9854 &  & 0.7500 & 0.8182 & 0.7826 \\
SED & LSTM & 0.9886 & 0.8529 & 0.9158 & CATSv2 & 0.4864 & 0.9727 & 0.6485 \\
(Out) & TimesNet & 0.8129 & 0.8541 & 0.8331 & (Out) & 0.4680 & 0.7137 & 0.5654\\
 & AT & 0.8234 & 0.7943 & 0.8087 &  & 0.5014 & 0.7083 & 0.5873\\
 & Timer & 0.7870 & 0.7596 & 0.7732 &  & 0.4984 & 0.7470 & 0.5980\\
 & Chronos & 0.8038 & 0.7981 & 0.8011 &  & 0.5032 & 0.7331 & 0.5969\\
 \rowcolor{gray!15} \cellcolor{white} & Qwen2.5-7B & 0.5714 & 1.0000 & 0.7273 & \cellcolor{white} & 0.4977 & 0.9818 & 0.6606 \\ \bottomrule
\end{tabular}}
\label{other-data-01clf}
\end{table}

\begin{table}[!t]
\centering
\caption{Performances of the recent algorithms and ~\dataname ~fine-tuned LLM for the anomaly detection.}
\renewcommand{\arraystretch}{1}
\resizebox{1\linewidth}{!}{
\begin{tabular}{l|c|ccc|ccc}
\toprule
\multirow{2}{*}{Training Type} & \multirow{2}{*}{Model} & \multicolumn{3}{c|}{\textbf{Univariate TS}}  & \multicolumn{3}{c}{\textbf{Multivariate TS}}  \\ 
\cline{3-8}
 &  & P & R & F1 & P & R & F1\\ 
\midrule
 & KNN & 0.9040 & 0.1129 & 0.2008 & 0.9019 & 0.1141 & 0.2026\\
Unsupervised& LOF & 0.9509 & 0.1161 &  0.2070 &0.9069  &  0.0967& 0.1748  \\
 Learning& AE1SVM & 0.9160 & 0.1177 & 0.2087  &0.8627 & 0.1091 & 0.1938\\ 
& DeepSVDD &  0.9165& 0.1141 & 0.2030 & 0.9298 & 0.1315 & 0.2304\\
\midrule
& XGBoost & 0.9074 & 0.9682 & 0.9368& 0.9162 & 0.9776 & 0.9459 \\
 & LightGBM & 0.9023 & 0.9758 & 0.9376 & 0.9316 & 0.9801 & 0.9552\\ 
Supervised& LSTM  & 0.8300 &0.9976  & 0.9061 & 0.7840 &  1.0000 &  0.8790\\
Learning& TimesNet  & 0.8521 & 0.9683 & 0.9066 & 0.8234 & 0.9890 & 0.8987 \\
& AT  & 0.8781 & 0.9702 & 0.9219 & 0.8823 & 0.9918 & 0.9339 \\
 \rowcolor{gray!15} \cellcolor{white} & Qwen2.5-7B & 0.8452 & 0.9295 & 0.8854 & 0.7557 & 0.9826 & 0.8544\\ \midrule

 & TimesFM & 0.7872 & 0.6258 & 0.6974 & 0.8435 & 0.7432 & 0.7903\\ 
Pre-trained & Timer & 0.8135 & 0.7295 & 0.7693 & 0.8257 & 0.7324 & 0.7764\\ 
Model& Chronos & 0.7976 & 0.7814 & 0.7895 & 0.8814 & 0.7945 & 0.8358\\
& MOMENT & 0.8005  & 0.6834 & 0.7374 & 0.8471 & 0.7775 & 0.8109\\
\rowcolor{gray!15}  \cellcolor{white} &  Qwen2.5-7B & 0.8465 & 0.9352 & 0.8886 & 0.7567 & 0.9801 & 0.8541\\
\bottomrule
\end{tabular}}
\label{other-alg-01clf-univar}
\end{table}

\textbf{RQ 4: Can fine-tuned LLMs ready for plug-and-play ~\modelname ~task?} We evaluate the fine-tuned \texttt{Qwen2.5-7B} models on real-world datasets from other domains and compare the results with those of common TSAD models. Tables~\ref{other-data-01clf} and~\ref{other-alg-01clf-univar} demonstrate that the fine-tuned \texttt{Qwen2.5-7B} possesses strong cross-domain transferability, effectively serving as a ``plug-and-play'' solution for the \modelname ~task. When applied directly to new datasets without additional tuning, \texttt{Qwen2.5-7B} consistently outperforms classic unsupervised methods like KNN, LOF, and AE1SVM\footnote{Evaluate by \href{https://github.com/yzhao062/pyod}{PyOD}. \label{pyod}}, and remains competitive with recent state-of-the-art deep learning and time series foundation models such as TimesNet, Anomaly Transformer (AT), Timer, and Chronos\footnote{Evaluate by \href{https://github.com/thuml/Time-Series-Library}{Time-Series-Library}.}. Notably, the model achieves exceptionally high recall across both in-domain (NEK, GHL) and out-of-domain (SED, CATSv2) datasets~\cite{liu2024elephant}, indicating a robust capability to detect anomalies in diverse real-world environments. While specialized supervised models like XGBoost\textsuperscript{\ref{pyod}} may maintain a lead in precision as shown in Table~\ref{other-alg-01clf-univar}, \texttt{Qwen2.5-7B} bridges the gap between traditional zero-shot approaches and supervised benchmarks. These results conclude that fine-tuned LLMs are ready for practical deployment in scenarios where labeled data is scarce, providing reliable diagnostic performance across different domains without the need for task-specific retraining.

\textbf{Further Analysis.} We provide qualitative case studies across diverse domains in Appendix~\ref{finetuning}, demonstrating the model's diagnostic logic for both univariate and multivariate anomalies. We also conduct a failure case analysis in Appendix~\ref{Failure—case}, identifying limitations in detecting subtle drifts or overlapping patterns. Furthermore, ablation studies in Appendix~\ref{Prompt-Design-Ablation-Study} show that prompt engineering, specifically few-shot examples and Chain-of-Thought reasoning, enhances reasoning coherence and categorization accuracy.

\section{Related Work}
Due to space limitations, we put the related work in Appendix~\ref{relatework}.

\section{Conclusion}
In this paper, we introduced Time-series Reasoning for Anomaly, \modelname, a new reasoning-focused anomaly detection task addressing critical gaps in traditional approaches by integrating detection, fine-grained classification, and causal explanation. To support this task, we constructed \dataname, ~the first real-world multimodal dataset with detailed high-quality annotations generated through structured prompting and \texttt{GPT-4} refinement. Extensive evaluations demonstrate that structured fine-tuning enhances anomaly detection and interpretability. Our work establishes a foundation for future research, underscoring the potential of generative, reasoning-enhanced anomaly detection models in real-world applications.

\section*{Limitations}
While \modelname~and the \dataname ~dataset facilitate progress in anomaly diagnosis, several limitations persist. Currently, the framework primarily identifies a single dominant anomaly category per sequence and may overlook secondary, co-occurring patterns. Furthermore, processing extremely long time-series sequences remains challenging due to LLM token constraints, and while visual representations provide a bridge, they do not fully replace the need for specialized long-context architectural adaptations. Finally, although fine-tuned models exhibit strong transferability, maintaining high diagnostic precision across vastly different domains without task-specific retraining remains a frontier for future exploration. Future work may extend \modelname ~to multi-label or hierarchical time series anomaly diagnosis.

\section*{Ethical Considerations}
The construction of \dataname ~adheres to ethical research standards, utilizing publicly available, open-source repositories where sensitive data (e.g., healthcare ECG) has been de-identified at the source. This study involves human subjects as annotators. We have provided all participants with clear instructions, the details of which are documented in the Section~\ref{data_quality_sec} and Appendix~\ref{data_quality} (recruitment, compensation, and data consent protocols were not applicable for this study). We emphasize that the diagnostic reasoning generated by our system is designed to serve as a decision-support tool for human experts, not as a replacement for professional judgment in high-stakes environments like medical diagnosis or industrial safety. Users should be mindful of potential model hallucinations in complex scenarios and are encouraged to deploy these tools responsibly to ensure human-in-the-loop verification. Overall, our approach does not involve any personally identifiable information (PII) or sensitive data, and we adhere to responsible AI practices by following the ACL Code of Ethics.

\section*{Acknowledgments}
We would like to thank the reviewers for their constructive feedback and insightful suggestions. We are also deeply grateful to the creators of the various open-source benchmarks, including, but not limited to, UCR, NASA, Yahoo, and others, whose foundational datasets and research made the development of the \dataname ~benchmark possible. We also want to thank those open-source repos and libraries for our evaluation and baseline comparison. (1) For classical and machine learning TSAD baseline comparison: \url{https://github.com/yzhao062/pyod}. (2) For deep learning and foundation model TSAD baseline comparison: \url{https://github.com/thuml/Time-Series-Library}. (3) For other TSAD datasets and descriptions: \url{https://github.com/TheDatumOrg/TSB-AD}. Your open-source contributions make it possible to conduct a fair and efficient comparison.

\bibliography{reference}

\newpage

\appendix

\begingroup
  \renewcommand{\contentsname}{Appendix Contents}
  \setcounter{tocdepth}{3} 
  \tableofcontents
\endgroup
\newpage

\section{Related Work}\label{relatework}
\textbf{Time Series Anomaly Detection}. Time series anomaly detection involves identifying patterns in temporal sequences that deviate significantly from expected normal behavior~\cite{blazquez2021review,boniol2024dive}. Early approaches relied on statistical methods (e.g., Z-score~\cite{chikodili2020outlier}, moving averages, exponential smoothing~\cite{phillips2021business}, and ARIMA~\cite{box1970distribution}) and decomposition-based methods (e.g., HP Filter and STL~\cite{gao2020robusttad,zhang2022tfad}). However, these methods often struggle with complex and non-stationary patterns~\cite{zamanzadeh2024deep}. ML-based methods~\cite{kant2019time,ruff2018deep,shin2020itad, karczmarek2020k,yang2021efficient} improved flexibility but typically require feature engineering or assumptions about the underlying anomaly patterns. DL-based methods, such as autoencoders~\cite{sakurada2014anomaly}, variational autoencoders~\cite{park2018multimodal}, and recurrent neural networks~\cite{su2019robust}, learn temporal dependencies and can capture complex nonlinearities. Transformer-based methods have further enhanced long-range dependency modeling and become state-of-the-art in many benchmark datasets~\cite{xu2021anomaly,yang2023dcdetector,zhong2024patchad}. Besides, approaches designed for time series foundation models, i.e., a unified ``one-fits-all'' framework, can also be adapted for anomaly detection~\cite{zhou2023one,wu2023timesnet,goswami2024moment,liang2024foundation}. Recently, there has been growing interest in applying LLMs or multimodal/vision-language models (MLLMs/VLMs) for the TSAD task~\cite{liu2024large2,zhou2024llmsunderstandtimeseries,xu2025can}.

Despite these advancements, challenges remain, including handling rare anomaly categories, ensuring robustness under noisy or missing data, and more detailed anomaly categories and attribution~\cite{zamanzadeh2024deep,jin2024position}. Recent works have started trying multimodal data fusion with novel datasets and LLM to address these issues~\cite{zhou2024llmsunderstandtimeseries,kong2025time}.

\textbf{Multimodal LLMs for Time Series Analysis}. LLMs/MLLMs have recently been explored and utilized for combining their powerful sequence modeling and reasoning abilities. As for the TSAD task, early studies show that LLM-based methods can achieve detection performance comparable to DL-based models and provide clear explanations~\cite{liu2024large2}. LLMAD enhances interpretability and performance by retrieving similar segments and applying anomaly-aware chain-of-thought prompting~\cite{liu2024large}. Others convert time series into images for VLMs to visually detect patterns and anomalies~\cite{zhou2024llmsunderstandtimeseries,xu2025can}. They concluded that LLMs often understand time series better when it’s presented visually rather than as text or numerical data. However, the same work noted that simply prompting LLMs to reason through the time series did not lead to significant performance gains~\cite{zhou2024llmsunderstandtimeseries}. For forecasting tasks, researchers~\cite{chattime2025aaai,kong2025time} adapt LLMs by formatting time series as text with relative information, allowing general LLMs to perform zero-shot or few-shot predictions.

Despite recent progress, the application of LLMs to time series analysis remains in its early stages and faces several key challenges. LLMs seem limited in detecting subtle or complex anomalies that are not explicitly covered in their training data or prompts, especially in real-world, context-dependent scenarios~\cite{zhou2024llmsunderstandtimeseries}. Moreover, effectively combining different data modalities in a unified model is non-trivial, and usually requires careful prompting based on the specifics of each task~\cite{kong2025position}, also lacks real datasets for model training and evaluation~\cite{xu2025can,liu2024large}. Therefore, we propose the \modelname~task that devises novel reasoning paradigms to improve the reliability of LLM-driven time series analysis.

\textbf{Data Resources in Anomaly Detection and Reasoning}. A range of public datasets supports research in the TSAD task. Single-modality benchmarks include univariate datasets (e.g., UCR~\cite{dau2019ucr} and Yahoo~\cite{laptev2015s5}) and multivariate datasets (e.g., NASA’s MSL and SMAP~\cite{hundman2018detecting} and NIPS-TS~\cite{lai2021revisiting}). Recently, time series-based multimodal datasets have emerged, combining temporal data with textual logs or images and attracting growing attention. Examples include the AIOps challenge dataset that integrates performance metrics, logs, and traces from microservices~\cite{bakhtin2025lo2}, and the Nonastreda dataset, pairing machine sensor time series with microscope images to detect manufacturing tool anomalies~\cite{truchan2025nonastreda}. There are also some general multimodality time series datasets from multiple application areas~\cite{kong2025position}. However, comprehensive benchmarks that incorporate numeric, textual, and visual modalities remain rare. Also, the existing limited multimodal data for time series reasoning for anomaly mostly comes from synthetic data, lacking real-world complexity~\cite{xu2025can,zhou2024llmsunderstandtimeseries}.

Notably, current datasets for time series reasoning and anomaly interpretation remain limited. Although synthetic datasets and tasks, such as Time-MQA (time series question answering)~\cite{kong2025time}, have been proposed to evaluate reasoning capabilities, establishing widely accepted real-world labeled datasets for the \modelname ~task remains an open research challenge~\cite{chow2024towards,jin2024position,kong2025position}. To fill this gap, we design the first real-world multimodal dataset \dataname~that moves beyond anomaly detection toward deeper analytical reasoning.

\section{Boundary and Further Direction}
For the boundaries, there are some worth noting: \textbf{(i) Detection of Multiple Anomaly Types in a Single Sequence.} In cases where multiple types of anomalies coexist within a single time series segment, our model tends to identify only the most salient or dominant anomaly type. While the reasoning module may mention secondary or other anomaly patterns, the primary classification output reflects the most prominent category. This limitation arises from the model's training objective, which emphasizes accuracy in identifying the most impactful anomaly rather than exhaustively listing all types. In particular, for our dataset, the time series length is 16-128, so the probability of multiple anomalies occurring simultaneously is not very high. \textbf{(ii) Multivariate Anomaly Detection Bias.} Our current approach to multivariate time series focuses primarily on capturing inter-variable relationships. However, we do not explicitly categorize univariate anomalies that may occur independently within each dimension for the multivariate experiment. Although these univariate anomalies are often mentioned in the LLM-generated reasoning, they are not directly reflected in the anomaly type classification, potentially omitting finer-grained insights for each variable. \textbf{(iii) Reliability of the \dataname ~Dataset.} During the expert verification process for the \dataname ~dataset, experts were allowed to select and rank multiple plausible anomaly types per instance. In our evaluation, if the model’s predicted anomaly type matched any of the top-ranked expert-annotated types, particularly the most important one, we considered the prediction correct. While this evaluation strategy increases robustness, it may mask subtle discrepancies between model output and expert intent, especially in borderline or ambiguous cases.

Several directions remain for exploration. \textbf{(i) Scalability to Longer Time Series.} As time series scales in length, directly inputting the entire sequence for LLM-based reasoning becomes challenging due to the token length constraints. Future work could explore time-series-specific embedding or a hierarchical summarization strategy to compactly represent long sequences while preserving critical temporal patterns. Alternatively, enhancing the weight of the accompanying visual representations (e.g., time series plots) may serve as a complementary approach. \textbf{(ii) Improved Modeling of Multiple Coexisting Anomalies.} Current classification is limited to predicting a single dominant anomaly type. Future research could focus on multi-label anomaly categorization that explicitly supports detection and reasoning over multiple overlapping anomaly types within the same sequence, potentially with structured output formats. \textbf{(iii) Continual Learning and Domain Adaptation.} Time series distribution may shift over time or across application domains (e.g., finance vs. Machine shown in Figure~\ref{fig:uni-case-study}). Developing continual learning frameworks or lightweight domain adaptation strategies compatible with reasoning models remains an important challenge for real-world deployment.

\section{The Definition of Anomaly Types}
To systematically address the challenge of time series anomaly detection, it is essential to first establish a clear and comprehensive taxonomy of anomaly types. Accordingly, we categorize anomalies into two primary classes: univariate and multivariate. These structured definitions of anomaly types are the foundational framework of \modelname~that guides the design of robust classification metrics.

Univariate anomalies, detailed in Table~\ref{Table:Uni-clf-des}, pertain to abnormality in a single time series. These can range from simple, isolated deviations such as \textit{Point Anomalies} and \textit{Sudden Spikes}, to more complex structural shifts. The classification encompasses changes in statistical properties (e.g., \textit{Change Point} and \textit{Distributional Change Anomaly}), alterations in long-term behavior (e.g., \textit{Trend Change} and \textit{Drift Anomaly}), and disruptions in cyclical or repetitive patterns (e.g., \textit{Periodic Change} and \textit{Pattern Change Anomaly}). Furthermore, we define anomalies of data acquisition failures, such as \textit{Sudden Flatline} and \textit{Repeated Value} anomalies. For each category, the table provides a formal definition, a representative visual illustration, and a practical domain example to facilitate intuitive understanding.

In contrast, multivariate anomalies arise from the complex interplay between variables. As cataloged in Table~\ref{Table:Mul-clf-des}, these anomalies are often subtle, as individual variables may appear normal when inspected in isolation. The abnormality lies in the violation of their expected joint behavior or inter-variable relationships. Our taxonomy covers several key types, including disruptions in statistical dependencies like the \textit{Covariance Structure Anomaly} and \textit{Collinearity Shift Anomaly}, and deviations in temporal synchronization, defined as \textit{Temporal Dependency Anomaly}. We also identify anomalies where variables diverge from a common pattern (\textit{Trend Divergence Anomaly}) or where their combined state is improbable (\textit{Joint Space Anomaly}). Finally, some anomalies are only detectable in a reduced-dimensional latent space (\textit{Principal Component Space Anomaly}). Similar to the univariate table, each multivariate anomaly type is accompanied by a precise definition, a two-variable visualization, and a real-world scenario to clarify its meaning.

\begin{table*}[!th]
\centering
\caption{Univariate anomaly types with example observations and explanation.}
\renewcommand{\arraystretch}{1.5}
\resizebox{\textwidth}{!}{
\begin{tabular}{m{4.2cm} m{5cm} m{4.5cm} m{4.5cm} }
\toprule
\textbf{Anomaly Category} & \textbf{Definition} & \textbf{Observation} & \textbf{Domain Example} \\
\midrule
Normal Sequence & There are no abnormal situations in this time series. & \includegraphics[width=\linewidth]{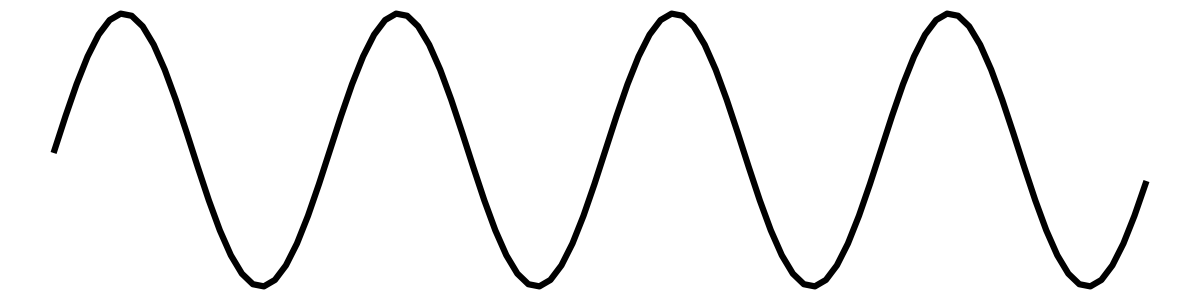} & Industrial sensor recording shows normal behavior \\
\midrule
Point Anomaly & A single data point significantly deviates from the local or global pattern. & \includegraphics[width=\linewidth]{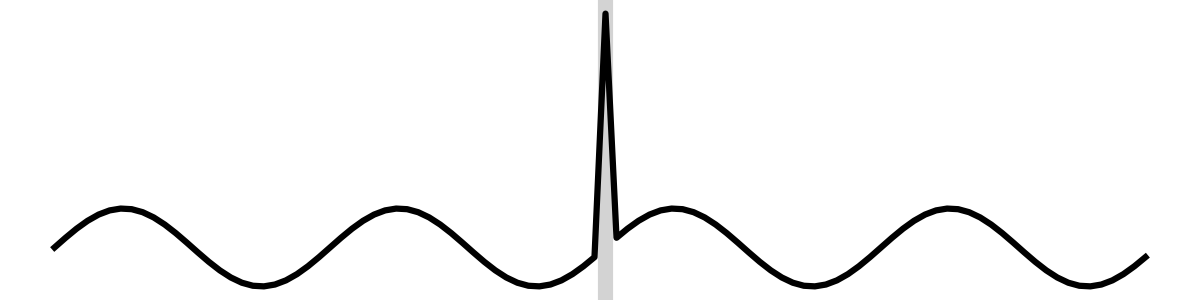} & Sudden spike in heart rate during sleep detected by a wearable device. \\
\midrule
Periodic Change Anomaly & The original periodic pattern is disrupted, such as broken periods or anomalous amplitude. & \includegraphics[width=\linewidth]{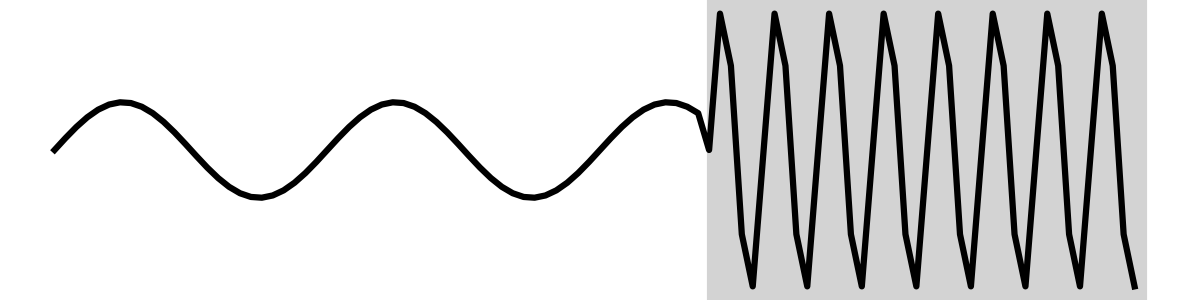} & Power consumption pattern changes due to a faulty machine affecting periodic load. \\
\midrule
Trend Change Anomaly & A sudden change in the long-term trend of the time series. & \includegraphics[width=\linewidth]{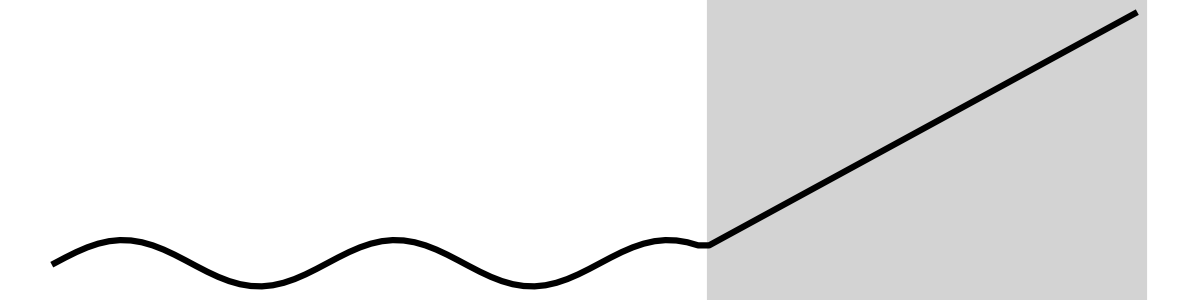} & Sales data shows a sudden increase due to a product launch. \\
\midrule
Change Point Anomaly & Statistical properties (e.g., mean, variance) change abruptly at certain points. & \includegraphics[width=\linewidth]{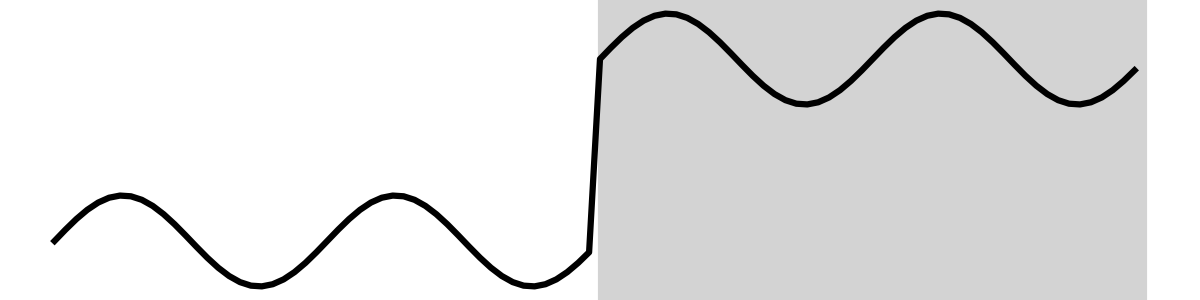} & Network latency shifts abruptly due to a routing change. \\
\midrule
Distributional Change Anomaly & The statistical distribution of the time series changes significantly. & \includegraphics[width=\linewidth]{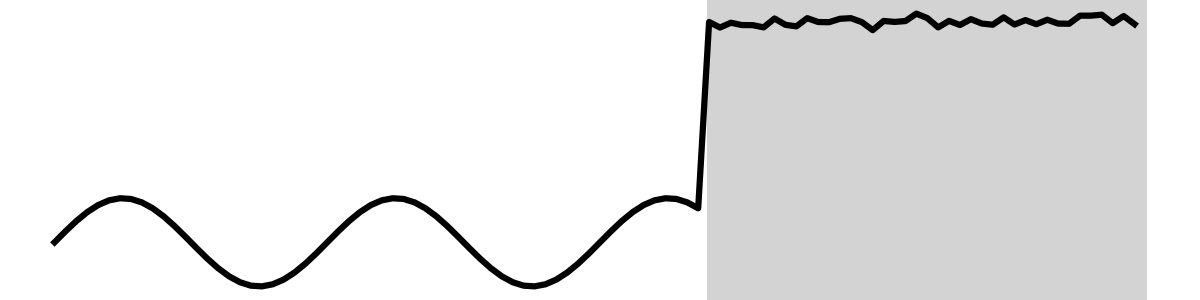} & A change in website traffic from normal to bot-generated patterns. \\
\midrule
Amplitude Anomaly & The amplitude of data points exceeds the normal upper or lower bounds. & \includegraphics[width=\linewidth]{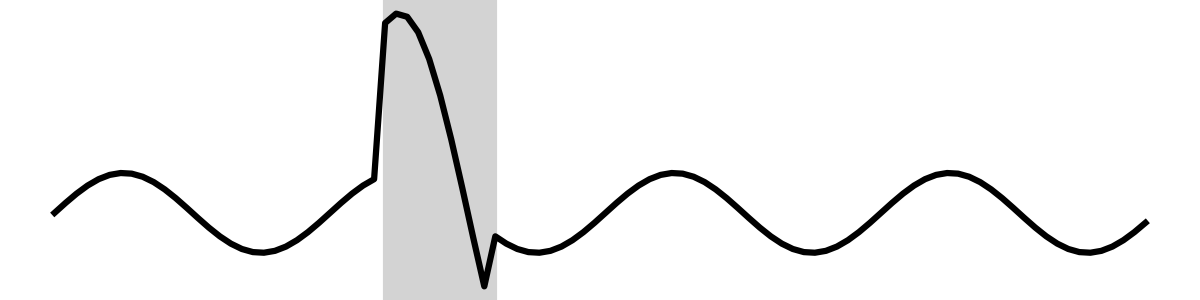} & ECG signal exceeds expected amplitude thresholds, indicating potential arrhythmia. \\
\midrule
Pattern Change Anomaly & The pattern of the time series suddenly changes from one form to another. & \includegraphics[width=\linewidth]{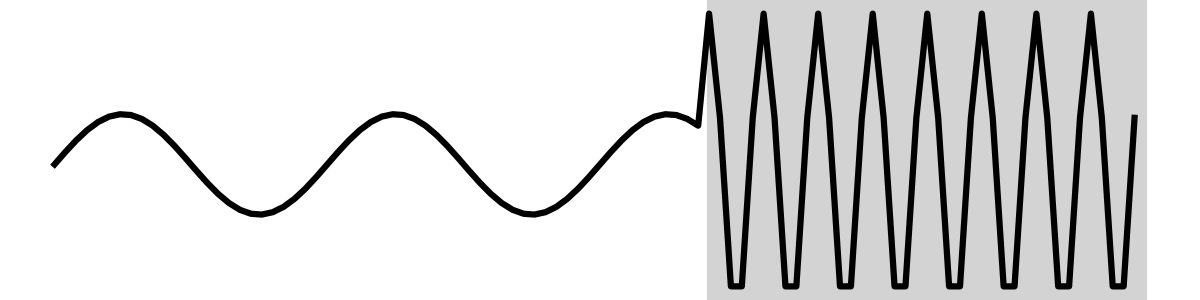} & Stock market index changes from oscillating to a steady downward trend. \\
\midrule
Sparse Anomaly & Isolated anomalous patterns occasionally appear in a long time series. & \includegraphics[width=\linewidth]{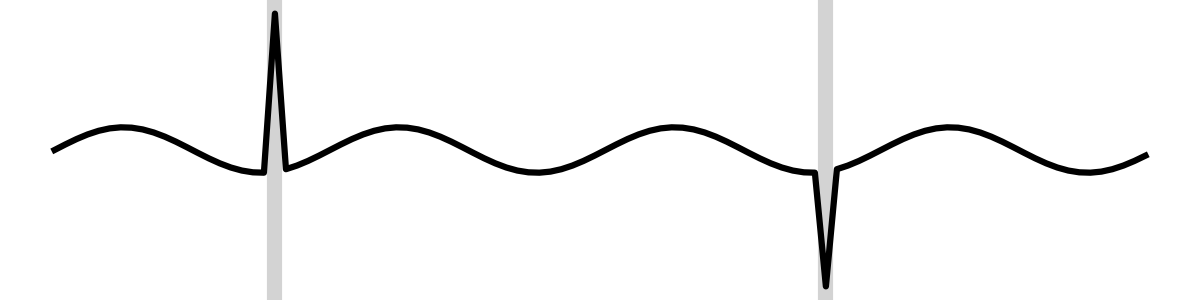} & Occasional fraudulent transactions in a large sequence of financial data. \\
\midrule
Repeated Value Anomaly & Continuous or intermittent repeated values disrupt the normal fluctuation pattern. & \includegraphics[width=\linewidth]{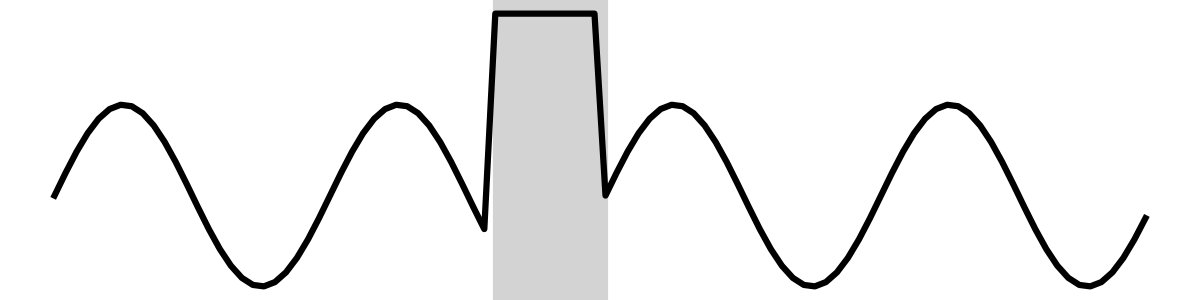} & Temperature sensor gets stuck and repeatedly reports the same value. \\
\midrule
Sudden Flatline Anomaly & The time series suddenly becomes a flat line with no normal fluctuations. & \includegraphics[width=\linewidth]{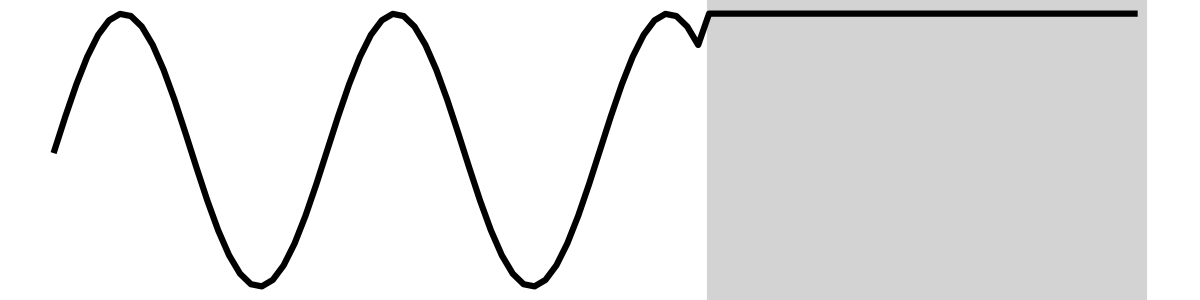} & IoT device disconnects and reports a flat signal for several minutes. \\
\midrule
Drift Anomaly & The data gradually drifts away from the normal level. & \includegraphics[width=\linewidth]{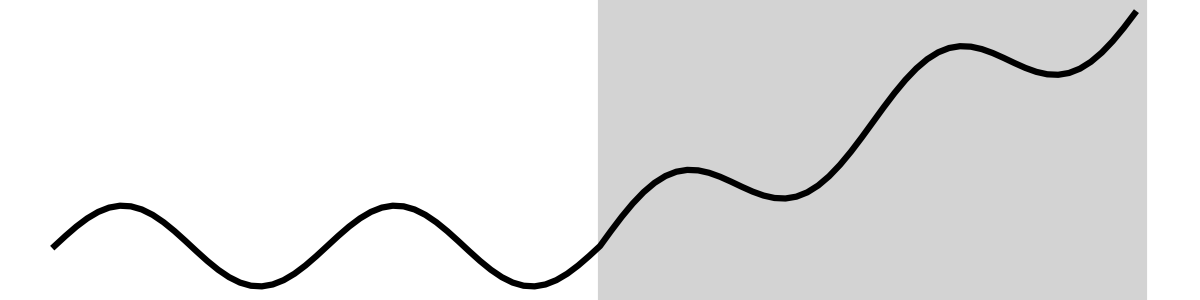} & GPS drift causes a slowly diverging location reading over time. \\
\midrule
Sudden Spike Anomaly & The data suddenly spikes or drops within a short time and then returns to normal. & \includegraphics[width=\linewidth]{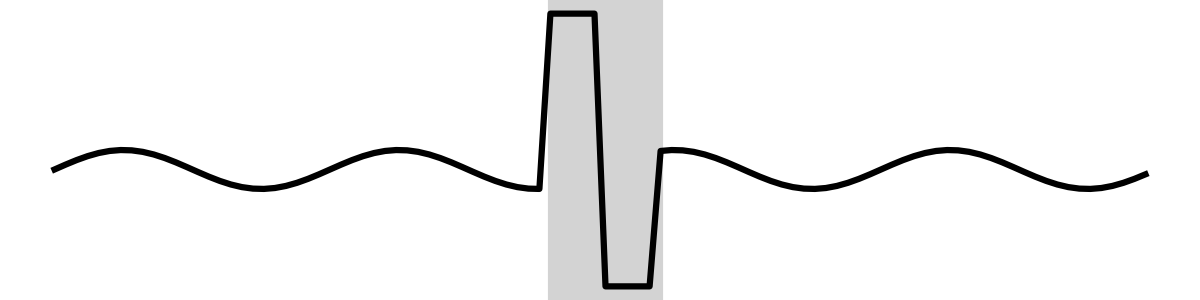} & Sudden voltage spike in an electrical grid followed by return to normal. \\
\midrule
Continuous Segment Anomaly & A continuous segment of data points deviates from the normal pattern. & \includegraphics[width=\linewidth]{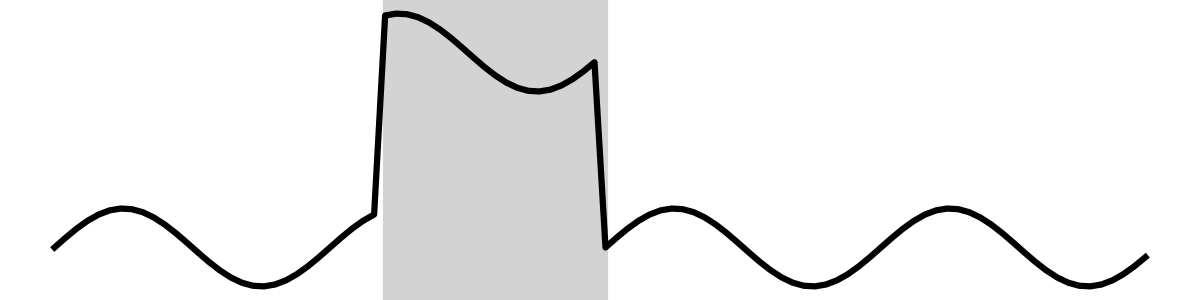} & A segment of network traffic deviates from expected behavior during a DDoS attack. \\
\midrule
Nonlinear Pattern Anomaly & Nonlinear changes appear, breaking the original linear rule. & \includegraphics[width=\linewidth]{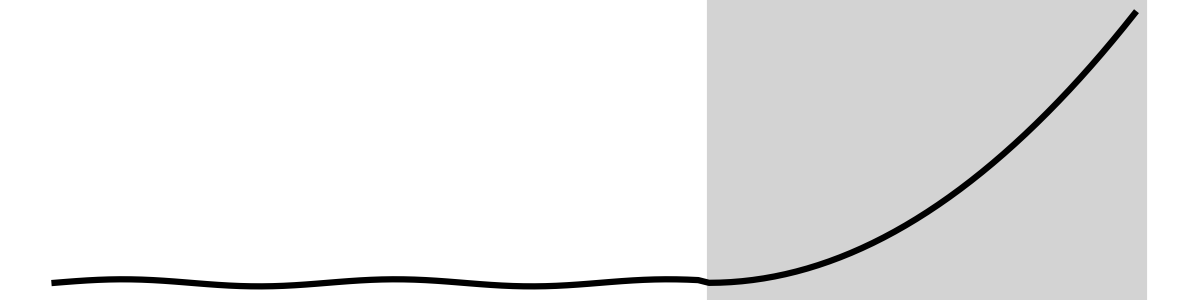} & A traffic speed pattern changes from linear increase to nonlinear surge due to congestion. \\
\bottomrule
\end{tabular}} 
\label{Table:Uni-clf-des}
\end{table*}

\begin{table*}[!th]
\centering
\caption{Multivariate anomaly types with example observations and explanation (two variables for example).}
\renewcommand{\arraystretch}{1}
\resizebox{\textwidth}{!}{
\begin{tabular}{m{2.8cm} m{6cm} m{4.2cm} m{5cm}}
\toprule
\textbf{Anomaly Category} & \textbf{Definition} & \textbf{Observation} & \textbf{Domain Example} \\
\midrule
Normal Sequence & From a multivariate view, all variables follow expected patterns over time. Relationships among variables and their dynamics remain stable without any abnormality. & \includegraphics[width=\linewidth]{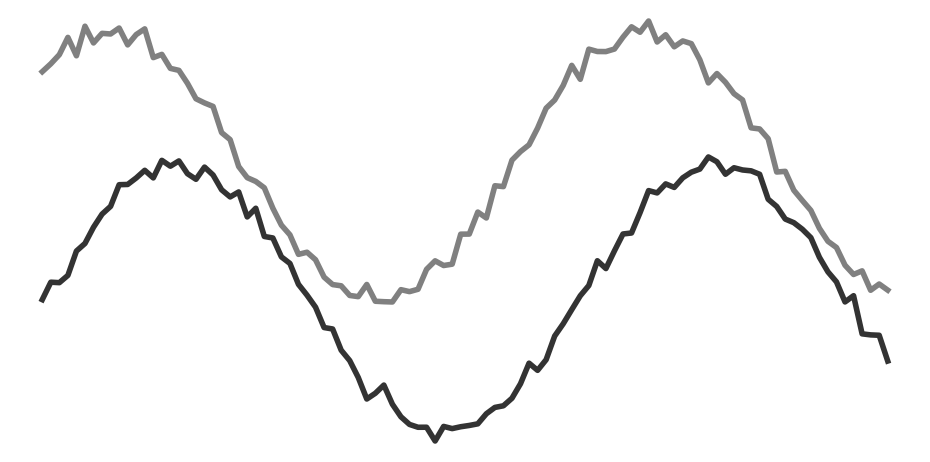} & A smart factory's sensors operate consistently: temperature, pressure, and vibration stay within expected ranges and show stable interdependencies. \\
\midrule
Covariance Structure Anomaly & The usual covariance or correlation structure among variables changes suddenly, such as reversal or unexpected decorrelation. & \includegraphics[width=\linewidth]{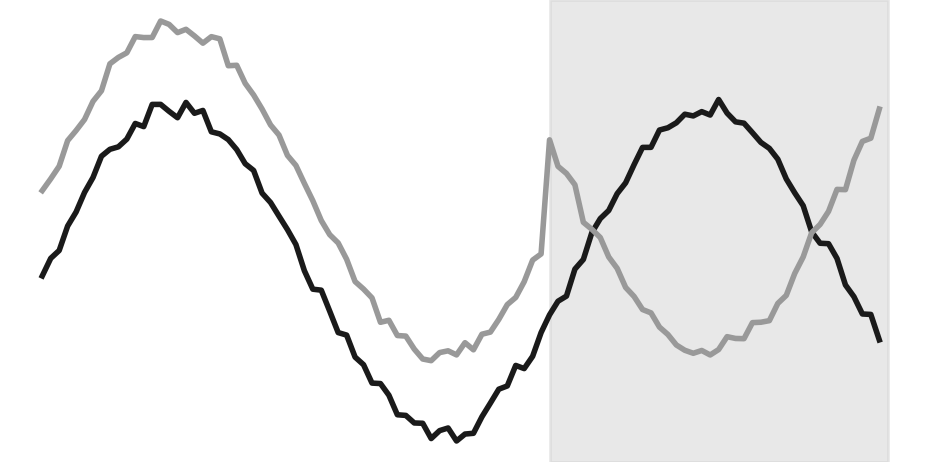} & In a financial system, the strong positive correlation between two stock prices suddenly breaks, potentially indicating market manipulation or systemic stress. \\
\midrule
Temporal Dependency Anomaly & Expected temporal dependencies (e.g., fixed lags and variable response delays) are violated, indicating possible desynchronization or timing failures. & \includegraphics[width=\linewidth]{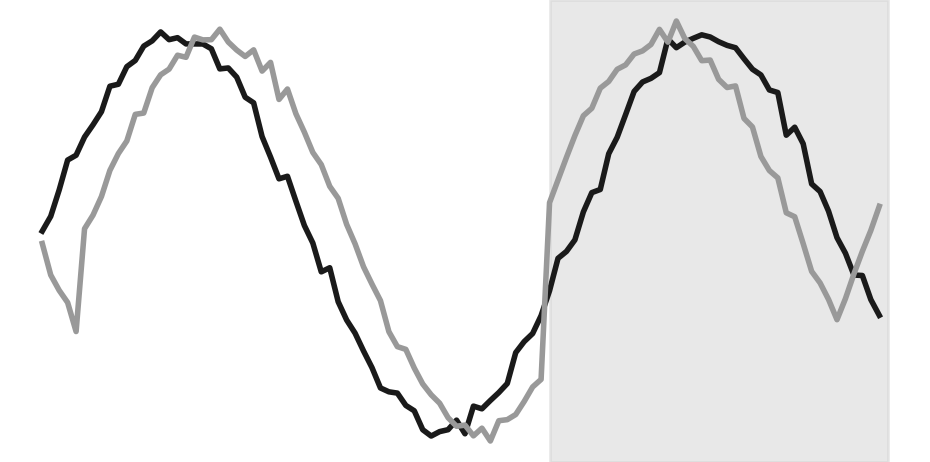} & In a manufacturing line, the normal delay between motor start and sensor response disappears, suggesting a sensor fault or a control system failure. \\
\midrule
Trend Divergence Anomaly & A subset of variables unexpectedly deviates from a shared trend, suggesting localized failures or partial system faults. & \includegraphics[width=\linewidth]{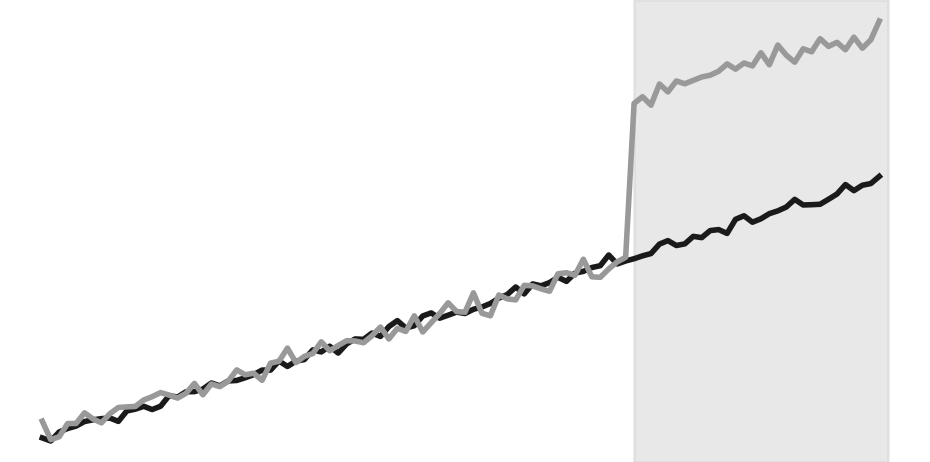} & In a power grid, one region’s voltage levels begin to drift from the national trend, possibly indicating equipment aging or a localized overload. \\
\midrule
Joint Space Anomaly & Although individual variable values may appear normal, their joint configuration is anomalous—suggesting system-level inconsistency in the multivariate space. & \includegraphics[width=\linewidth]{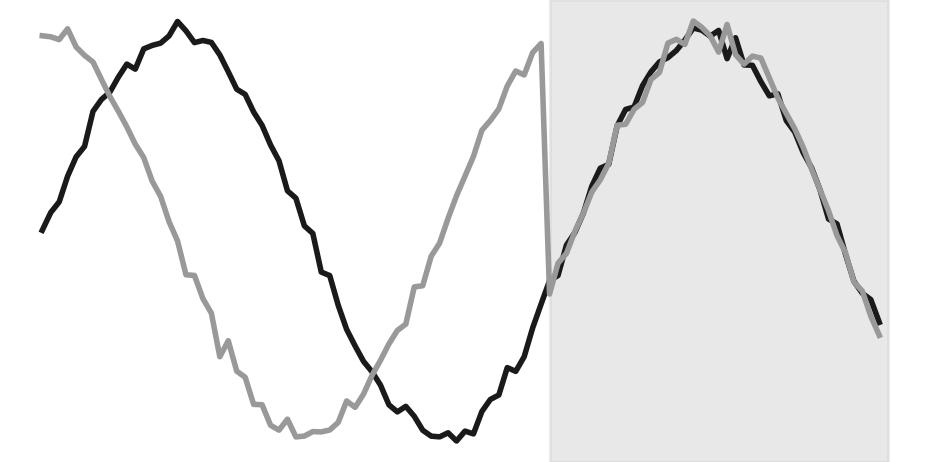} & In autonomous driving, speed and steering angle are each within normal limits, but their combination implies unsafe turning behavior. \\
\midrule
Principal Component Space Anomaly & An anomaly becomes evident only in a lower-dimensional latent space (e.g., PCA), revealing subtle structural deviation across many variables. & \includegraphics[width=\linewidth]{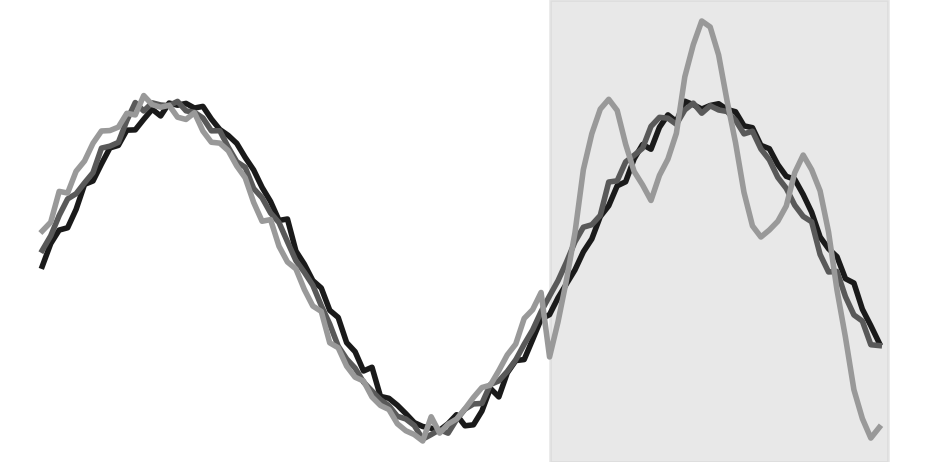} & In climate modeling, subtle but coordinated changes across dozens of climate indicators (e.g., temperature, pressure, humidity) show up only in the PCA space, signaling early signs of climate shifts. \\
\midrule
Collinearity Shift Anomaly & Strong linear dependencies or redundancies between variables suddenly break down, often due to malfunctioning or desynchronized components. & \includegraphics[width=\linewidth]{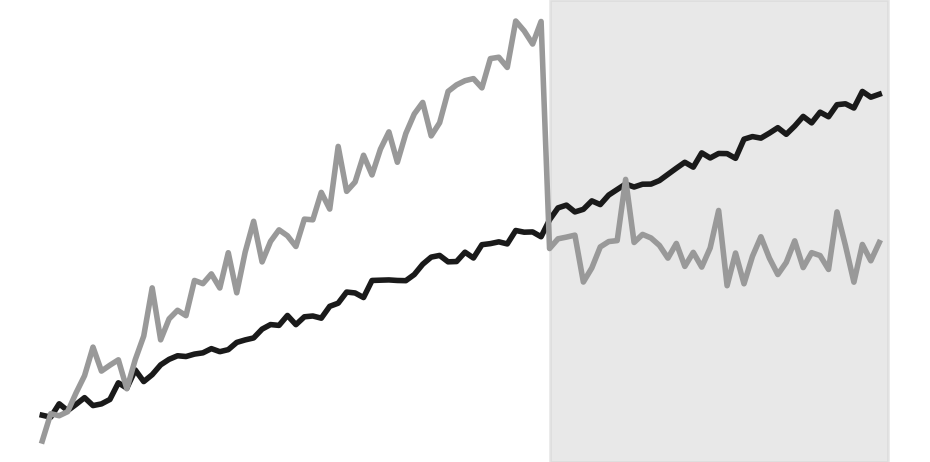} & In server monitoring, CPU and memory usage used to be tightly coupled, but suddenly became independent—suggesting possible memory leakage or process isolation failure. \\
\bottomrule
\end{tabular}}
\label{Table:Mul-clf-des}
\end{table*}

\section{Example of \modelname}\label{finetuning}
\begin{figure*}[!t]
    \centering
    \includegraphics[width=1\linewidth]{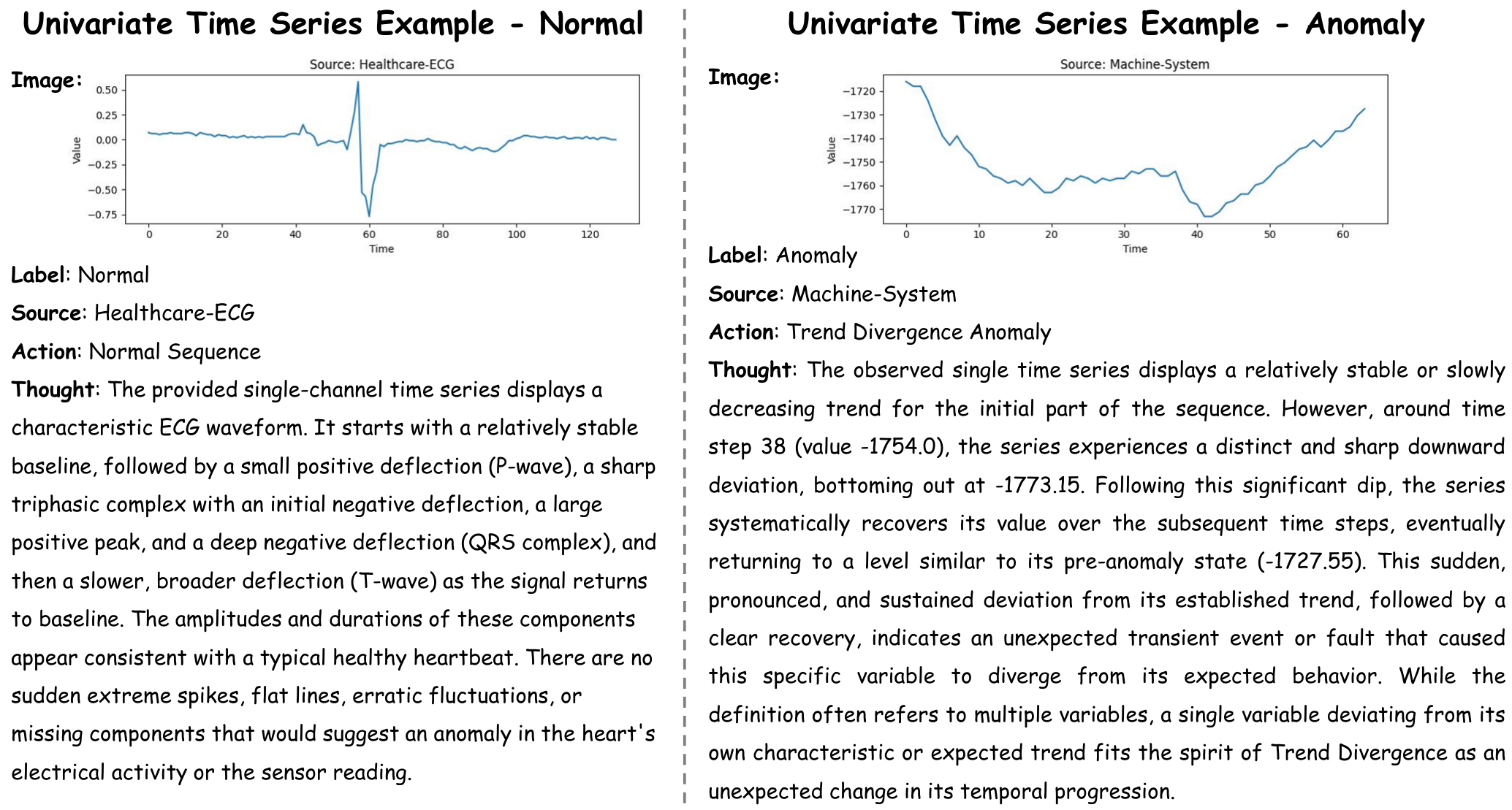}
    \caption{Univariate \modelname ~case study.}
    \label{fig:uni-case-study}
\end{figure*}

\begin{figure*}[!t]
    \centering
    \includegraphics[width=1\linewidth]{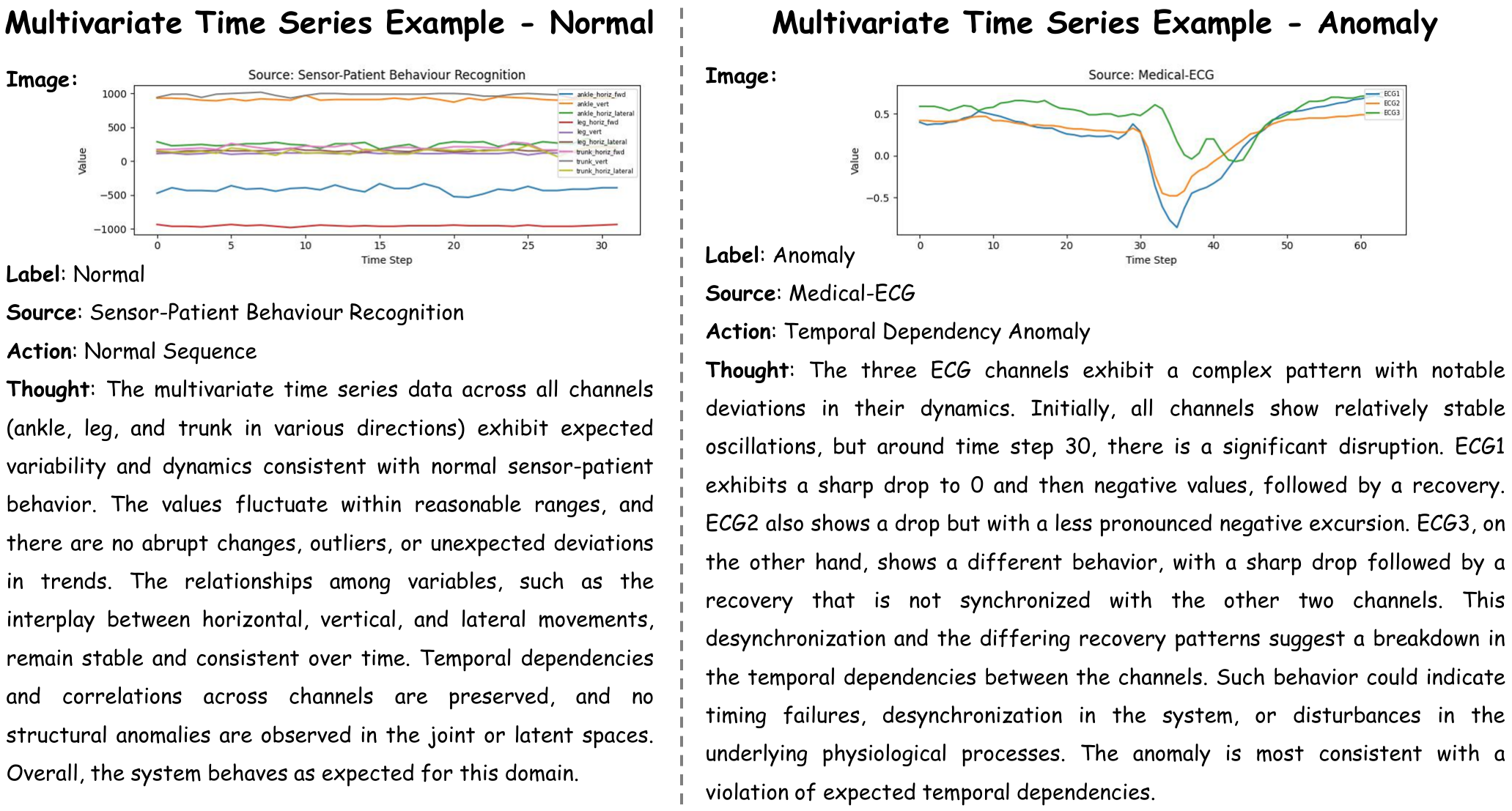}
    \caption{Multivariate \modelname ~case study.}
    \label{fig:multi-case-study}
\end{figure*}

This section demonstrates how SFT works for the \modelname ~task through two case studies. Table \ref{detail_univariate} shows univariate anomaly detection on ECG data, where the model identifies a nonlinear pattern anomaly through step-by-step reasoning. Table \ref{detail_multivariate} presents a multivariate case detecting temporal dependency anomalies across synchronized ECG channels. Supporting visualizations of these cases are shown in Figures \ref{fig:uni-case-study} and \ref{fig:multi-case-study}, illustrating the model's analytical process and decision-making. In the univariate case (Figure \ref{fig:uni-case-study}), source information is particularly important for accurate action classification and for grounding the relevance of the Thought. In the univariate case, such as the ``normal'' ECG example, source context is crucial for accurate classification and domain-specific Thought generation. Without it, healthcare terms may be omitted or anomalies misclassified as amplitude anomaly. Similarly, in multivariate cases (Figure \ref{fig:multi-case-study}), source and variable data provide essential context for interpreting Action and grounding Thought. In summary, Action can be correctly interpreted using the available multimodal data, while Thought can rely on both source and variable information for domain-specific attribution and precise anomaly localization.

\begin{table*}[h!]
\centering
\caption{An example of SFT for a univariate time series.}
\vspace{-3mm}
\label{detail_univariate}
\begin{tabular}{>{\raggedright\arraybackslash}p{0.95\textwidth}}
\hline
\#\#\textbf{User:}\\
You are an expert in time series anomaly detection. We provide a time series (called Observation), you should give us the anomaly type (called Action) and its reasons (called Thought). 
Thought steps can infer the current abnormal situation of a time series.\\
The anomaly detection of each time series is divided into three steps: Observation, Thought, and Action. After analyzing each observation, please provide the next Thought and next Action.
Here is a time series observation that we need to check for anomaly categories. The observation is from the domain of Healthcare-ECG.\\
Please make a Thought judgment and put your final Action within \textbackslash\textbackslash boxed1\{\} and \textbackslash\textbackslash boxed2\{\} respectively, where action must just be a category name not id.\\
\textbf{Observation}: -0.04, -0.05, -0.09, 0.0, 0.32, 0.37, -0.46, -1.56, -1.52, -0.98, -0.51, -0.2, -0.03, 0.0, -0.02, 0.1, 0.02, 0.11, 0.07, 0.13, 0.1, 0.24, 0.15, 0.14, 0.17, 0.24, 0.2, 0.2, 0.21, 0.23, 0.37, 0.24, 0.19, 0.22, 0.16, 0.16, 0.14, 0.07, 0.12, -0.01, 0.06, -0.06, 0.14, 0.03, 0.09, 0.07, -0.01, 0.06, 0.08, 0.03, 0.0, 0.0, -0.02, 0.09, 0.08, 0.1, 0.0, 0.05, 0.16, 0.09, 0.17, 0.13, 0.06, 0.16, 0.14, 0.11, 0.09, 0.19, 0.07, 0.13, 0.06, 0.09, -0.09, -0.18, -0.26, -0.06, 0.46, 1.41, 2.73, 4.34, 4.65, 4.34, 3.18, 2.55, 2.2, 2.87, 2.35, 1.57, 0.51, 0.05, -0.34, -0.48, -0.54, -0.54, -0.66, -0.81, -0.8, -0.97, -0.95, -1.09, -1.21, -1.53, -1.55, -1.68, -1.86, -2.09, -2.08, -2.22, -2.09, -2.05, -2.08, -1.67, -1.53, -1.28, -0.99, -0.82, -0.62, -0.39, -0.22, -0.11, -0.19, 0.06, -0.03, -0.05, 0.09, 0.03, -0.05, 0.1\\
\textbf{Thought}: \textbackslash\textbackslash boxed1\{\}\\
\textbf{Action}: \textbackslash\textbackslash boxed2\{\}\\
\\

\#\#\textbf{Assistant:}\\
\textbf{Thought}: \textbackslash\textbackslash boxed1\{The time series exhibits multiple anomalies. Initially, there is a sudden drop and recovery in values (e.g., -0.46 to -1.56 and back to near 0), followed by a gradual drift upward (e.g., from 0.46 to 4.65). 
This is then followed by a nonlinear pattern of exponential decay (e.g., from 4.65 to -2.22) and oscillations. The sequence shows a combination of drift, nonlinear changes, and potential periodic disruptions, which are characteristic of ECG anomalies. The most prominent anomaly is the nonlinear pattern of exponential growth and decay, which breaks the expected linear or periodic behavior of an ECG signal.\}\\
\textbf{Action}: \textbackslash\textbackslash boxed2\{Nonlinear Pattern Anomaly\}\\
\hline
\end{tabular}
\end{table*}

\begin{table*}[h!]
\centering
\caption{An example of SFT for a multivariate time series.}\vspace{-3mm}
\label{detail_multivariate}
\begin{tabular}{>{\raggedright\arraybackslash}p{0.95\textwidth}}
\hline
\#\#\textbf{User:}\\
You are an expert in time series anomaly detection. We provide a time series (called Observation), you should give us the anomaly type (called Action) and its reasons (called Thought). 
Thought steps can infer the current abnormal situation of a time series.\\

The anomaly detection of each time series is divided into three steps: Observation, Thought, and Action. After analyzing each observation, please provide the next Thought and next Action.
Here is a time series observation that we need to check for anomaly categories. The observation is from the domain of Medical-ECG.\\
Please make a Thought judgment and put your final Action within \textbackslash\textbackslash boxed1\{\} and \textbackslash\textbackslash boxed2\{\} respectively, where action must just be a category name not id.\\
\textbf{Observation}: "ECG1": [0.12, 0.11, 0.1, 0.07, 0.06, 0.04, 0.02, 0.02, 0.01, 0.01, 0.0, -0.01, -0.01, -0.01, 0.02, 0.09, 0.06, -0.26, -0.56, -0.71, -0.79, -0.85, -0.78, -0.62, -0.57, -0.54, -0.53, -0.52, -0.42, -0.3, -0.19, -0.09, -0.02, 0.05, 0.14, 0.23, 0.26, 0.27, 0.27, 0.28, 0.29, 0.3, 0.3, 0.32, 0.34, 0.34, 0.36, 0.38, 0.39, 0.41, 0.41, 0.43, 0.43, 0.42, 0.42, 0.4, 0.37, 0.34, 0.3, 0.26, 0.2, 0.16, 0.14, 0.09]; "ECG2": [0.06, 0.06, 0.06, 0.05, 0.05, 0.04, 0.02, 0.01, 0.01, 0.01, 0.0, -0.01, -0.01, -0.01, -0.01, 0.03, 0.0, -0.15, -0.53, -0.79, -0.81, -0.76, -0.7, -0.58, -0.49, -0.45, -0.39, -0.33, -0.28, -0.18, -0.12, -0.06, -0.01, 0.04, 0.1, 0.16, 0.2, 0.2, 0.2, 0.2, 0.21, 0.21, 0.21, 0.21, 0.23, 0.23, 0.24, 0.24, 0.24, 0.26, 0.26, 0.26, 0.27, 0.27, 0.26, 0.24, 0.23, 0.2, 0.17, 0.13, 0.1, 0.07, 0.04, 0.02]; "ECG3": [0.14, 0.14, 0.14, 0.09, 0.07, 0.06, 0.04, 0.03, 0.01, 0.01, 0.01, -0.01, -0.01, 0.0, 0.0, 0.0, 0.01, 0.05, 0.07, 0.06, -0.06, -0.07, -0.1, -0.1, -0.08, 0.03, 0.01, -0.12, -0.28, -0.38, -0.45, -0.39, -0.26, -0.1, -0.01, 0.01, 0.03, 0.1, 0.13, 0.11, 0.14, 0.13, 0.14, 0.14, 0.14, 0.16, 0.17, 0.17, 0.2, 0.21, 0.2, 0.2, 0.23, 0.24, 0.25, 0.23, 0.22, 0.24, 0.23, 0.2, 0.2, 0.18, 0.12, 0.07]\\
\textbf{Thought}: \textbackslash\textbackslash boxed1\{\}\\
\textbf{Action}: \textbackslash\textbackslash boxed2\{\}\\
\\
\#\#\textbf{Assistant:}\\
\textbf{Thought}: \textbackslash\textbackslash boxed1\{The three ECG channels exhibit a shared general structure, but there are notable deviations in their dynamics and relationships. Initially, all channels show a gradual decline, followed by a sharp negative deflection (e.g., around time steps 17–21). However, the magnitude and timing of this deflection differ across channels. For example, ECG1 shows a more pronounced and earlier negative peak compared to ECG2 and ECG3. Additionally, ECG3 exhibits an unusual behavior around time steps 26–30, where it deviates significantly from the patterns of ECG1 and ECG2, showing a delayed and less synchronized recovery. This desynchronization and violation of expected temporal dependencies between the channels suggest a potential anomaly in the timing or coordination of the system, possibly due to sensor misalignment, physiological irregularities, or external interference.\}\\
\textbf{Action}: \textbackslash\textbackslash boxed2\{Temporal Dependency Anomaly\}\\
\hline
\end{tabular}
\end{table*}

\section{Statistics of the Model Pool}\label{llm-com-section}
\begin{figure*}[t]
    \centering
    \subfloat[Anomaly rate (\textcolor{blue}{blue}) and thought length (\textcolor{orange}{orange}) for different models.\label{fig:anly_fig}]{%
        \includegraphics[width=0.42\textwidth]{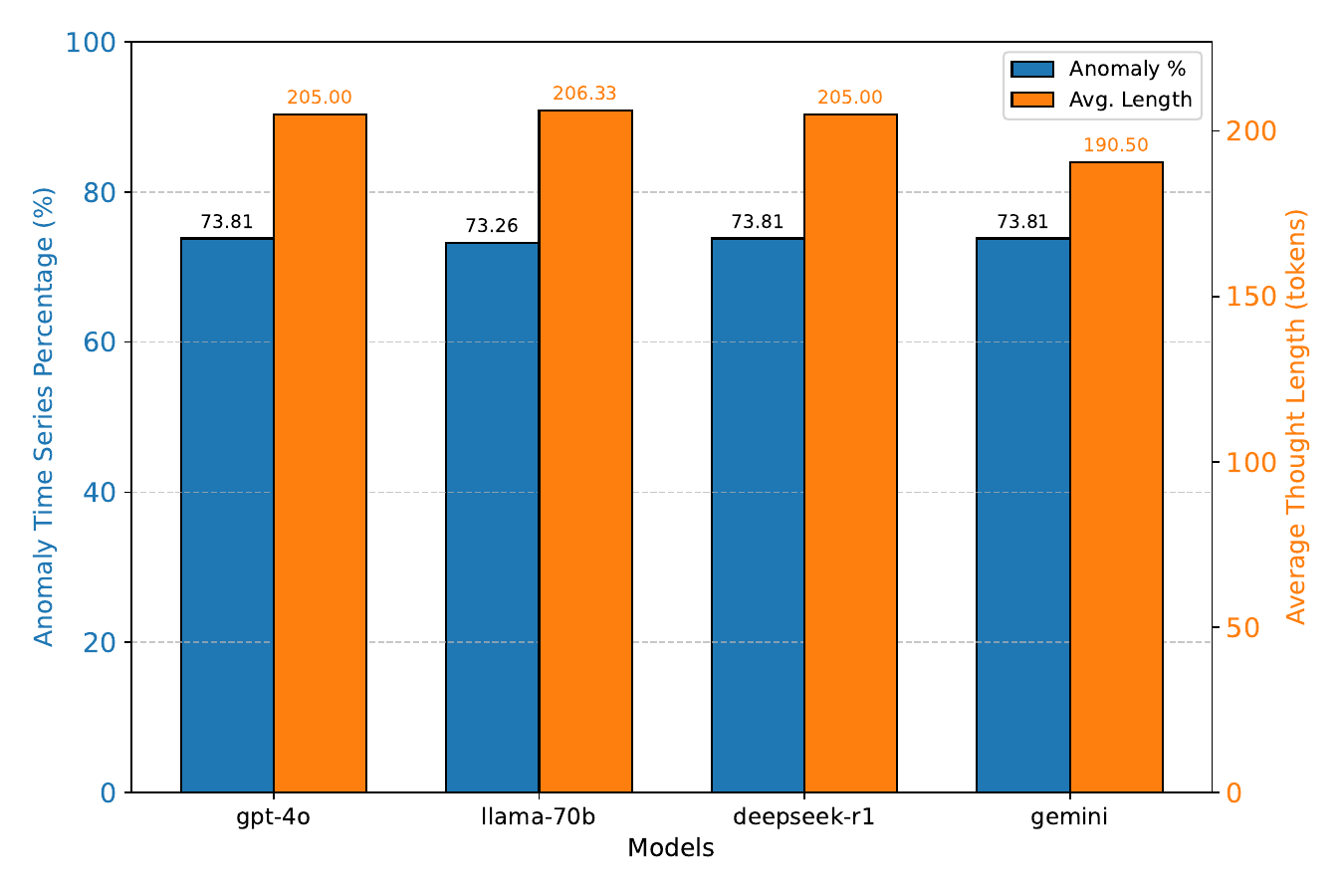}%
    }\hfill
    \subfloat[Classification of anomalies in different models with univariate and multivariate time series.\label{fig:data_fig}]{%
        \includegraphics[width=0.57\textwidth]{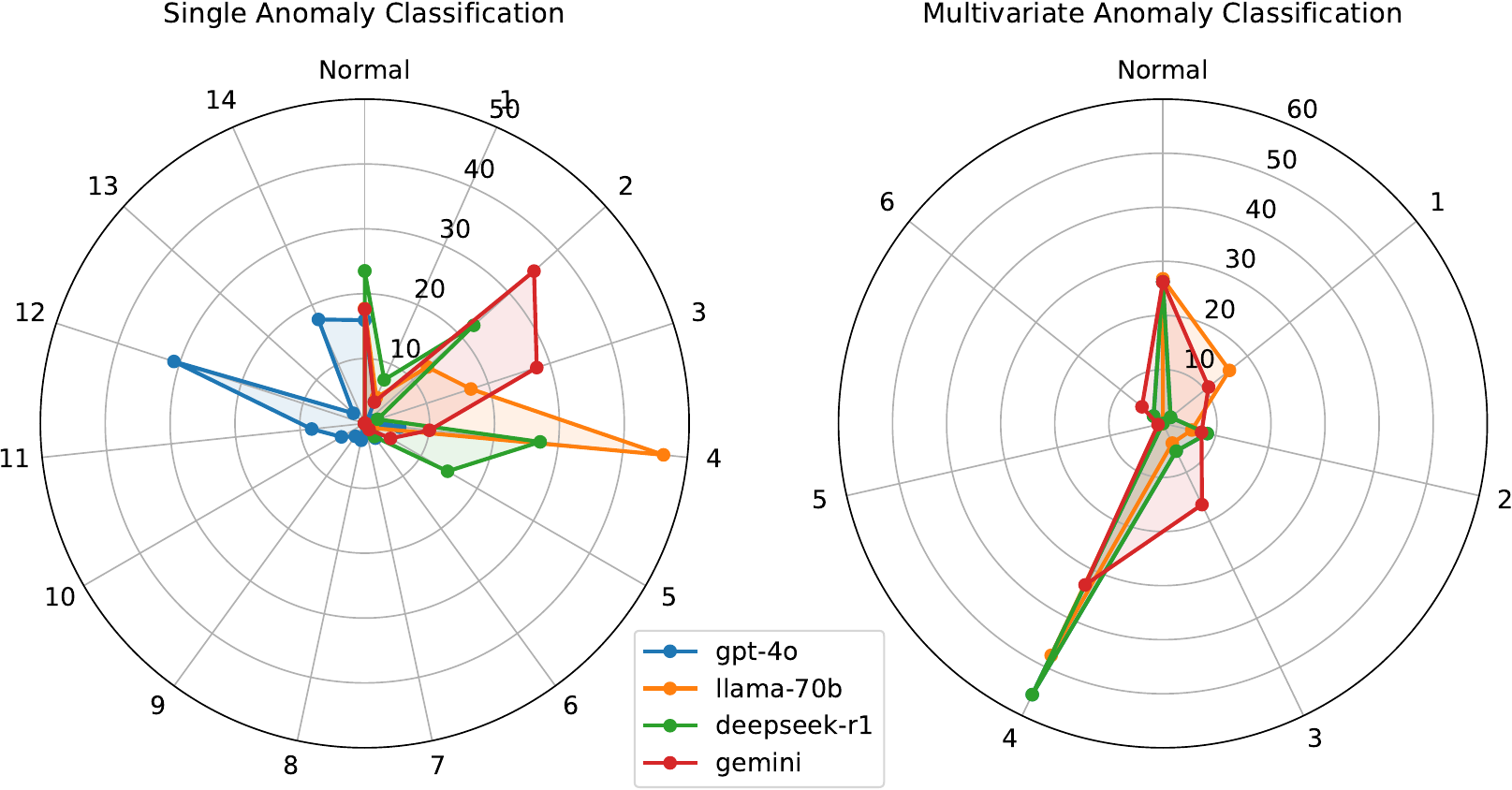}
    }
    \caption{Global statistics of model response in the model pool.}
    \label{fig:llm-comparison}
\end{figure*}

We further analyse the statistics of model responses in Figure~\ref{fig:llm-comparison}. Since we take as input the label of whether or not it is anomalous, the four strong models have similar anomaly judgements and follow the instruction label. However, as shown in Figure~\ref{fig:data_fig}, they differed in their performance on the task of classification of the source of anomalies, especially for the univariate time series. Further leading to differences in the length of the anomaly reasons in Figure~\ref{fig:anly_fig}.

\begin{table*}[ht]
\caption{Summary of models and links used in our experiments.}
\setlength{\tabcolsep}{3.5pt}
\centering
\begin{tabular}{l|l}
\toprule
\textbf{Model Name} & \textbf{Link}  \\
\midrule
DeepSeek-7B  & \url{https://huggingface.co/deepseek-ai/deepseek-llm-7b-chat} \\
Llama-3-8B &  \url{https://huggingface.co/meta-llama/Meta-Llama-3-8B-Instruct} \\
Llama-3.2-3B  & \url{https://huggingface.co/meta-llama/Llama-3.2-3B-Instruct} \\
Phi-4-mini &  \url{https://huggingface.co/microsoft/Phi-4-mini-instruct} \\
Qwen2.5-3B & \url{https://huggingface.co/Qwen/Qwen2.5-3B-Instruct} \\
Qwen2.5-7B & \url{https://huggingface.co/Qwen/Qwen2.5-7B-Instruct} \\
\midrule
Llava-v1.5-7B  & \url{https://huggingface.co/llava-hf/llava-1.5-7b-hf} \\
Llava-v1.5-13B   & \url{https://huggingface.co/llava-hf/llava-1.5-13b-hf} \\
Llama-3.2-11B-v  & \url{https://huggingface.co/meta-llama/Llama-3.2-11B-Vision}\\
Qwen2.5-vl-7B  & \url{https://huggingface.co/Qwen/Qwen2.5-VL-7B-Instruct}\\
\midrule
Gemini-2.5-flash & \url{https://deepmind.google/models/gemini/flash/}\\
DeepSeek-R1 & \url{https://huggingface.co/deepseek-ai/DeepSeek-R1}\\
Llama-3.3-70B-Instruct  & \url{https://huggingface.co/meta-llama/Llama-3.3-70B-Instruct}\\
gpt-4o\_2024-11-20  & \url{https://platform.openai.com/docs/models/gpt-4o}\\
GPT-4  & \url{https://platform.openai.com/docs/models/gpt-4}\\
\bottomrule
\end{tabular}
\label{llms}
\end{table*}

\begin{table*}[!ht]
\centering
\caption{Interpretation of quality scores for time series reasoning for anomaly.}
\begin{tabular}{c|p{10cm}}
\toprule
\textbf{Score} & \textbf{Interpretation} \\
\midrule
1 & Very Poor – Invalid, irrelevant, or nonsensical \\
\midrule
2 & Poor – Major issues or weak alignment \\
\midrule
3 & Fair – Partially reasonable but lacking clarity or depth \\
\midrule
4 & Good – Mostly clear, relevant, and logical \\
\midrule
5 & Excellent – Accurate, specific, useful, and well-written \\
\bottomrule
\end{tabular}

\label{tab:explanation_score}
\end{table*}

\begin{table*}[!th]
\centering
\caption{Evaluation dimensions and criteria for time series reasoning for anomaly.}
\resizebox{\textwidth}{!}{
\begin{tabular}{p{3.5cm}|p{6.5cm}|p{6.5cm}}
\toprule
\textbf{Dimension} & \textbf{Description} & \textbf{What to Look For} \\
\midrule
\textbf{Language Quality} & Is the reason grammatically correct, clearly phrased, and structurally complete? The sentence should read fluently, without ambiguity or major grammatical issues. & (1) Proper sentence structure (2) No grammatical or syntactic errors (3) Clear and fluent wording \\
\midrule
\textbf{Factual Soundness} & Is the reasoning factually aligned with the time series behavior (e.g., trend, spike, drop) and contextual data? The explanation should be plausible and consistent with the actual data. & (1) Matches visible pattern in the time series (2) Correctly reflects contextual data (3) Avoids hallucinations or random guesses \\
\midrule
\textbf{Specificity to Anomaly} & Does the reason specifically address the anomaly point in question? It should not be vague, overly generic, or a templated explanation that could apply to any anomaly. & (1) Mentions a concrete possible cause for a certain anomaly (2) Avoids boilerplate text (3) Shows evidence of localization in time or context \\
\midrule
\textbf{Interpretability} & Is the reason easy for a human to understand and follow? It should be logically structured, free from jargon, and ideally follow a cause-effect or descriptive reasoning pattern. & (1) Logical flow of thought (2) Minimal technical jargon (3) Easily understood by analysts or domain experts \\
\midrule
\textbf{Usefulness} & Does the explanation provide actionable insight or support further steps such as labeling, filtering, or alerting? A useful reason helps humans make better decisions. & (1) Supports human labeling decisions (2) Offers insight that could influence next steps (3) Helps explain or validate the anomaly for stakeholders \\
\bottomrule
\end{tabular}}

\label{tab:anomaly_explanation_criteria}
\end{table*}

\section{LLM Descriptions}~\label{LLM_describe}
Throughout this paper, the experiments use the version of off-the-shelf models detailed in Table~\ref{llms}. All models are selected in their instruction-tuned variants to ensure they can effectively follow prompts and perform the specified tasks. To provide a comprehensive analysis, we organize our selection into three distinct categories: open-source LLMs, VLMs, and state-of-the-art models accessed via APIs.

\section{Likert Scale for Evaluating LLM-Generated Annotation} \label{data_quality}
To assess the quality of the automatically generated reasoning, we design a structured human evaluation protocol based on a Likert scale. Specifically, we evaluate each explanation along five key dimensions: language quality, factual soundness, specificity to the anomaly, interpretability, and usefulness for downstream tasks. Table~\ref{tab:anomaly_explanation_criteria} defines each dimension along with practical indicators. Table~\ref{tab:explanation_score} describes the interpretation of Likert scores from 1 (very poor) to 5 (excellent).

\section{Generation and Annotation Templates}\label{Annotation}
Our data generation and refinement process is governed by a series of structured templates. The initial \textbf{Instruction for Anomaly Detection} Template prompts a model to perform time series anomaly detection, generating a Thought (reasoning) and Action (classification) for a given observation. We then use a \textbf{Annotation Template} to have a judge model, such as \texttt{GPT-4}, score and rank the outputs from multiple LLMs in the model pool based on a predefined 5-point rubric. Finally, a \textbf{GPT-4 Critique Feedback} Template instructs an expert model to review and, if necessary, rewrite a model's response to create a gold-standard data sample, which is then used for further training. This multi-stage process ensures the creation of a high-quality, well-annotated dataset.

\section{The Details of the \dataname ~Dataset} \label{details_dataset}
The table~\ref{Dataset_details} lists how many time series segments were extracted from each subdataset. Specifically, we select raw time series from comprehensive open source repositories across diverse domains, including AIOps systems~\cite{laptev2015s5,liu2024elephant}, environment~\cite{barrenetxea2019sensorscope,liu2024elephant}, finance~\cite{ahmad2017unsupervised,tran2016distance}, healthcare~\cite{wu2021current,bachlin2009wearable,goldberger2000physiobank,greenwald1990improved}, IoT~\cite{barrenetxea2019sensorscope,ahmad2017unsupervised}, industrial sensors~\cite{wu2021current,liu2024elephant}, server data~\cite{ahmad2017unsupervised,liu2024elephant}, traffic~\cite{ahmad2017unsupervised}, network records~\cite{ahmad2017unsupervised,jacob2020exathlon}, and synthetic dataset~\cite{thill2020markusthill,lai2021revisiting}. Each subdataset provides at least 200 samples, with more than 80\% of them being anomalous. The sampling imbalance mainly arises from differences in the original subdataset sizes and lengths, as well as inherent differences in anomaly ratios across domains. For example, some of the Healthcare datasets are derived from ECG signals. Although the data length is similar to that of Finance datasets, the anomaly ratio in ECG is higher, leading to imbalanced sampling. Empirically, fine-tuning large models can partially generalize across such sampling imbalances. Moreover, we include the source information as part of our prompt, which can potentially mitigate the effects of imbalance.

Regarding the UCR datasets, although there are 250 subsequences, we treat each subsequence independently. A segment is labeled as anomalous if between 0 and 80\% of its original timestamps are annotated as anomalies within a given sampling window. We acknowledge that some segments may contain overlapping timestamps, but we ensure that no two segments are identical. In addition, for each subdataset, we set multiple segment lengths to further increase the number of final samples.

As shown in Listing~\ref{lst:Dataset}, a typical item based on a time series segment is structured in the \dataname ~dataset. The meaning of each key is as follows: ``Action'' is the anomaly category, ``ActionID'' is the number corresponding to the anomaly category used for numerical evaluation, ``Figurepath'' is the image corresponding to the time series segment, ``Observation'' corresponds to the original time series itself (if it is multivariate, we provide the variable name and a brief explanation), ``Source'' indicates the data source and specific domain, and ``Thought'' is the ground truth of the reasoning.

\begin{tcolorbox}[colback=white,colframe=gray!50!black,title=Instruction for Anomaly Detection, breakable]
\footnotesize
\textbf{Univariate Time Series Instruction}: You are an expert in time series anomaly detection. We provide a time series (called Observation), you should give us the anomaly type (called Action) and its reasons (called Thought). Thought steps can infer the current abnormal situation of a time series. Action is an abnormal category with the following 0-14 types, where 0 is a normal category. The explanations of 0-14 actions are as follows:\\
\{UNIVARIATE TIME SERIES ANOMALY CATEGORY\}\\

\textbf{Multivariate Time Series Instruction}: You are an expert in multivariate time series anomaly detection. We provide a multivariate time series (called Observation), where each time point contains multiple variables. Your task is to identify the anomaly type (called Action) and provide detailed reasoning (called Thought). The Thought should analyze the relationships, dynamics, and structures across all variables and time points to infer any abnormal behavior. The Action must be one of the following seven types, where type 0 means no anomaly. The definitions are:\\
\{MULTIPLE TIME SERIES ANOMALY CATEGORY\}
\tcblower
\footnotesize
The anomaly detection of each time series is divided into three steps: Observation, Thought, and Action. After analyzing each observation, please provide the next Thought and next Action. Here are some examples:\\
\{SAMPLES OF EXPERT DEFINITIONS\}\\

Here is a univariate/multivariate time series observation that we need to check for anomaly categories. We already know that it is a \{ANOMALY LABEL\} sequence and from the domain of \{SAMPLE SOURCE\}.
Please make a Thought judgment within \textbackslash\textbackslash boxed1\{\} and put your final Action in \textbackslash\textbackslash boxed2\{\} respectively, where action must just be a category name, not id. \\
Observation: \{SAMPLE OBSERVATION\} \\
Thought: \textbackslash\textbackslash boxed1\{\} \\
Action: \textbackslash\textbackslash boxed2\{\}
\end{tcolorbox}

\begin{tcolorbox}[colback=white,colframe=gray!50!black,title=Annotation Template, breakable]
\footnotesize
A task we have is:\\
$[$\{EXAMPLE INSTRUCTION FOR ANOMALY DETECTION\}$]$

Now we have the outputs of models, there are: \\
The model [gpt-4o] output is:\\
Thought: \{GPT-4O THOUGHT\} \\
Action: \{GPT-4O ACTION\}\\

The model [llama-70b] output is:\\
Thought: \{LLAMA-70B THOUGHT\} \\
Action: \{LLAMA-70B ACTION\}\\

The model [deepseek-r1] output is:\\
Thought: \{DEEPSEEK-R1 THOUGHT\} \\
Action: \{DEEPSEEK-R1 ACTION\}\\

The model [GEMINI] output is:\\
Thought: \{GEMINI THOUGHT\} \\
Action: \{GEMINI ACTION\}\\

Please evaluate the consistency between the output of each model and the task intent, and score and provide reasons for the answers of each model. The score is from 1 to 5:\\
1. **Irrelevant**: No alignment.\\
2. **Partial Focus**: Poor handling in a certain aspect, such as misclassification of exceptions.\\
3. **Partial Compliance**: The classification of anomalies is accurate, but there may be slight deviations or neglect of others in the reasons.\\
4. **Almost There**: Alignment close to expert answers, slight deviation.\\
5. **Comprehensive Compliance**: Completely consistent with expert answers, meeting all requirements.\\

Based on the above ratings, please provide me with a ranking to compare the output results from the above models. 

Here is an example of the output format:\\
<|begin|>gpt-4o>gemini>deepseek-r1>llama-70b<|end|>
\end{tcolorbox}

\begin{tcolorbox}[colback=white,colframe=gray!50!black,title=GPT-4  Critique Feedback, breakable]\label{gpt4cr}
\footnotesize
A task we have is:\\
$[$\{EXAMPLE INSTRUCTION FOR ANOMALY DETECTION\}$]$

Given the model answer to an instruction, your role is to provide specific and constructive feedback for me. When you review the model answer, consider its helpfulness, truthfulness, honesty, and how well it followed the instructions.

The model answer is:\\
Thought: \{THOUGHT\} \\
Action: \{ACTION\}\\

I need you to assume the role of an anomaly detection expert. It's essential that your feedback not only highlights areas for improvement but also provides actionable suggestions to help the model understand how to enhance its performance. Please make improvements based on the thought and action of the model and follow the same output. If no improvement is needed, just return **None**.

The following are examples of formats that need to be improved for output:\\
Thought: \{GPT-4 THOUGHT\} \\
Action: \{GPT-4 ACTION\}
\end{tcolorbox}

\begin{table*}[t]
\centering
\caption{The details of the proposed \dataname ~dataset.}
\label{tab:datasets}
\resizebox{0.73\textwidth}{!}{
\begin{tabular}{l|l|l|l}
\toprule
\textbf{Type} & \textbf{Dataset} & \textbf{Number of Segments} & \textbf{Key of Source} \\
\midrule
Univariate 
& TODS and NAB-artificial & 220   & Synthetic data \\
& NAB-Cloudwatch          & 277   & AIOps \\
& MBA-ECG and MITDB       & 14,394 & Healthcare \\
& NAB-Exchange            & 254   & Finance \\
& IOPS\_KPI               & 3,171  & Server \\
& MGAB                    & 900   & IoT \\
& SensorScope             & 4,776  & Environment \\
& NAB-Traffic             & 220   & Traffic \\
& NAB-Tweets              & 582   & Network \\
& UCR                     & 2,358  & Industrial sensors \\
& YAHOO                   & 9,148  & Server \\
\midrule
Multivariate
& GECCO                   & 914   & IoT \\
& Daphnet                 & 360   & Industrial sensors \\
& TAO                     & 1,000  & Environment \\
& MITDB, SVDB, and LTDB   & 1,000  & Healthcare \\
\bottomrule
\end{tabular}}
\label{Dataset_details}
\end{table*}

\begin{lstlisting}[language=json, caption={A time series segment example in RATs40K dataset}, label={lst:Dataset}]
{
  "Action": "Joint Space Anomaly",
  "ActionID": 4,
  "FigurePath": "figures_multi/train/27.jpg",
  "Label": "Anomaly",
  "Observation": {
    "ECG1": "[Real time series data]",
    "ECG2": "[Real time series data]",
    "ECG3": "[Real time series data]"
  },
  "Source": "Medical-ECG",
  "Thought": "The reasoning text"
}
\end{lstlisting}

\begin{table*}[ht]
\centering
\caption{Results of the prompt design ablation study. We report the performance of a fine-tuned \texttt{Qwen2.5-7B} model on univariate time series.}
\resizebox{\textwidth}{!}{
\begin{tabular}{ccc|ccc|ccc|cccc}
\toprule
\multirow{2}{*}{\textbf{Prompt}} 
& \multirow{2}{*}{\textbf{\# Shots}} 
& \multirow{2}{*}{\textbf{CoT}} 
& \multicolumn{3}{c|}{\textbf{Label Matching}} 
& \multicolumn{3}{c|}{\textbf{ActionID Matching}} 
& \multicolumn{4}{c}{\textbf{Thought Matching}} \\

\cline{4-6} \cline{7-9} \cline{10-13}
& & 
& P & R & F1 
& P & R & F1 
& Cosine & TFIDF & Lev. & Token \\
\midrule

Base & 0 & -- 
& 0.8205 & 0.9280 & 0.8710 
& 0.1132 & 0.1108 & 0.0743 
& 0.8683 & 0.2129 & 0.1783 & 0.1003 \\

\rowcolor{gray!8}
Base & 0 & \checkmark 
& 0.8474 & 0.9273 & 0.8857 
& 0.1206 & 0.1019 & 0.0767 
& 0.8812 & 0.2381 & 0.1820 & 0.1036 \\

Base & 1 & -- 
& 0.8199 & 0.9295 & 0.8714 
& 0.1490 & 0.1152 & 0.0802 
& 0.8927 & 0.2367 & 0.1873 & 0.1063 \\

\rowcolor{gray!8}
Base & 1 & \checkmark 
& 0.8485 & 0.9159 & 0.8810 
& 0.1543 & 0.1108 & 0.0815 
& 0.9090 & 0.2518 & 0.1929 & 0.1101 \\

Base & 7 (half) & -- 
& 0.8252 & 0.9297 & 0.8744 
& 0.1736 & 0.1013 & 0.0909 
& 0.9107 & 0.2717 & 0.2232 & 0.1388 \\

\rowcolor{gray!8}
Base & 7 (half) & \checkmark 
& 0.8463 & 0.9323 & 0.8873 
& 0.1739 & 0.1042 & 0.0918 
& 0.9165 & 0.3008 & 0.2390 & 0.1591 \\

\midrule
Simplified & 14 (full) & -- 
& 0.8218 & 0.9785 & 0.8934 
& 0.1916 & 0.1292 & 0.0959 
& 0.9117 & 0.2725 & 0.2298 & 0.1468 \\

\rowcolor{gray!8}
Simplified & 14 (full) & \checkmark 
& 0.8371 & 0.9643 & 0.8963 
& 0.1989 & 0.1486 & 0.0991 
& 0.9208 & 0.2948 & 0.2401 & 0.1593 \\

Base & 14 (full) & -- 
& 0.8286 & 0.9310 & 0.8769 
& 0.2498 & 0.1081 & 0.1051 
& 0.9198 & 0.2857 & 0.2328 & 0.1519 \\

\rowcolor{gray!8}
Base & 14 (full) & \checkmark 
& 0.8452 & 0.9295 & 0.8854 
& 0.2532 & 0.1229 & 0.1100 
& 0.9270 & 0.3160 & 0.2432 & 0.1624 \\

\bottomrule
\end{tabular}}
\label{Prompt_AB_study}
\end{table*}

\section{Prompt Design Ablation Study} \label{Prompt-Design-Ablation-Study}

In addition to the ablation study of image and observation shown in Table~\ref{tab_pass1-acc} of the main text, we conduct further detailed ablation studies to analyze the effect of prompt design on model performance, focusing on prompt format, the number of in-context examples (zero-, one-, and 7-shot), prompt variants, and the inclusion of chain-of-thought (CoT) reasoning. Table~\ref{Prompt_AB_study} shows that increasing the number of few-shot examples leads to consistent improvements, especially for anomaly category classification and reasoning-related metrics, while gains on label matching are comparatively modest. Adding chain-of-thought prompting provides additional benefits across most settings, with the most noticeable improvements observed in thought matching metrics, suggesting more structured and coherent reasoning outputs. At higher shot numbers (e.g., 14-shot), the performance differences between prompt structures become smaller, indicating that sufficient in-context examples can partially compensate for prompt complexity.

\begin{figure*}[!t]
    \centering
    \includegraphics[width=1\linewidth]{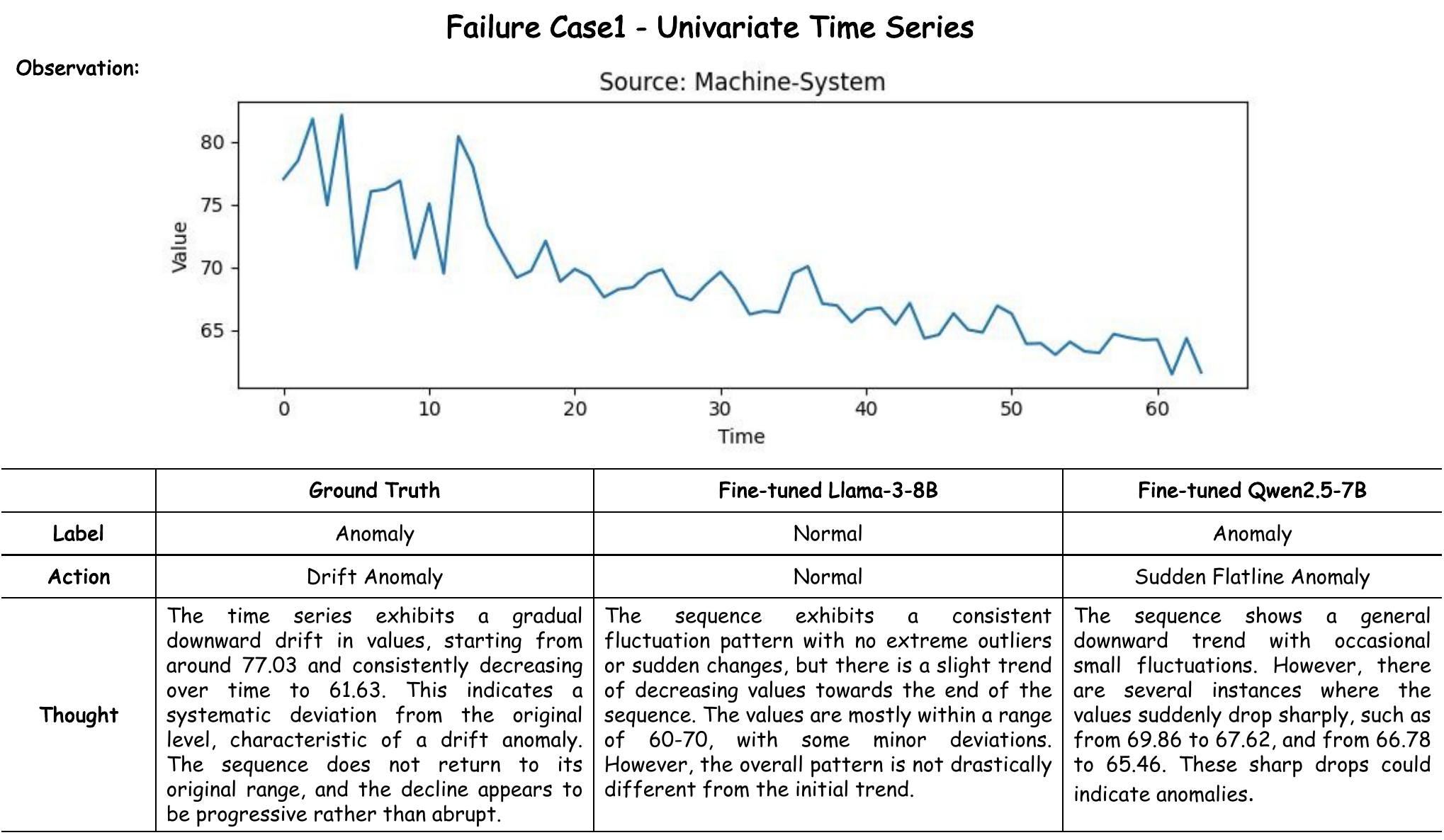}
    \caption{Univariate time series failure case 1.}
    \label{fig:failure_example1}
\end{figure*}

\begin{figure*}[!t]
    \centering
    \includegraphics[width=1\linewidth]{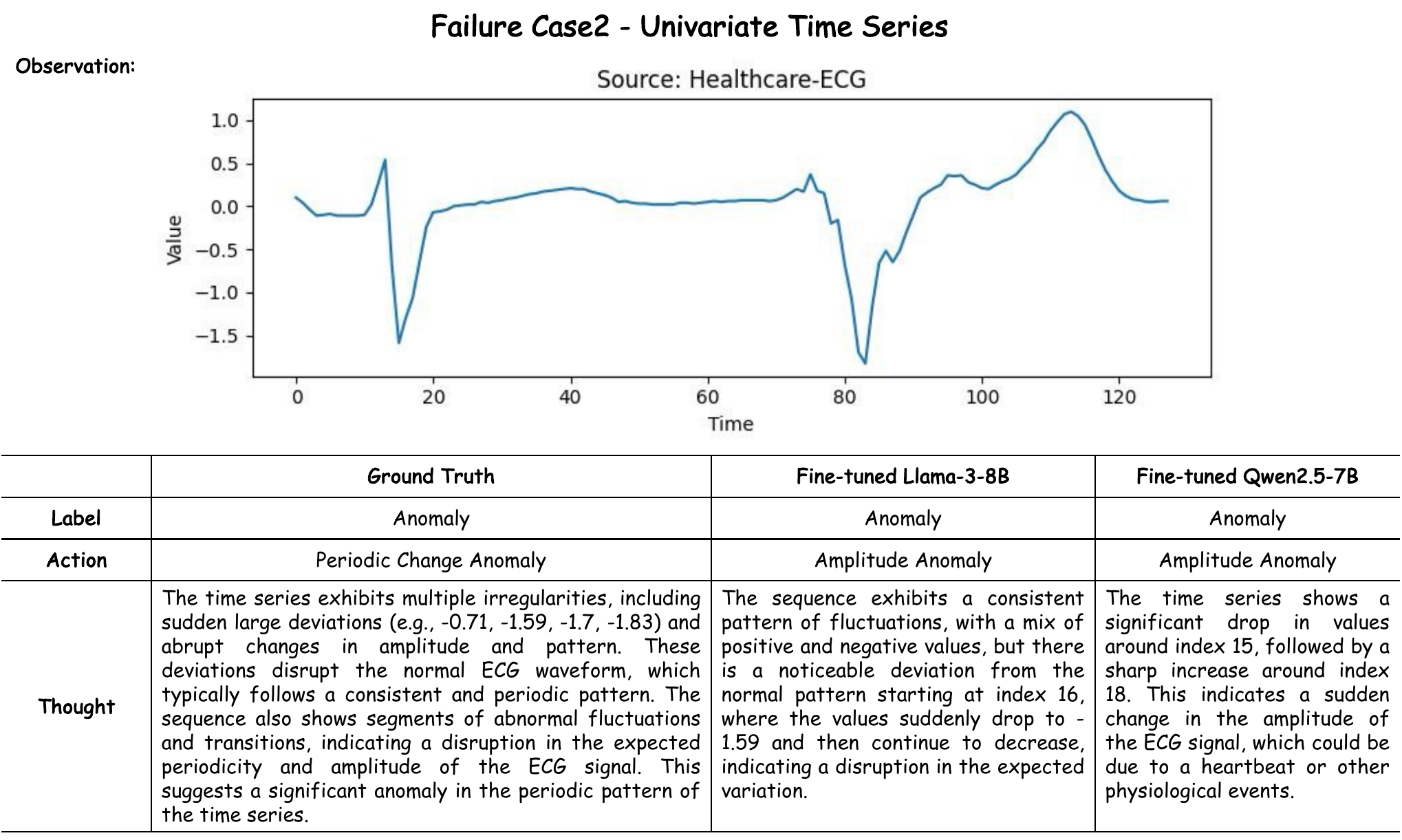}
    \caption{Univariate time series failure case 2.}
    \label{fig:failure_example2}
\end{figure*}

\begin{figure*}[!t]
    \centering
    \includegraphics[width=1\linewidth]{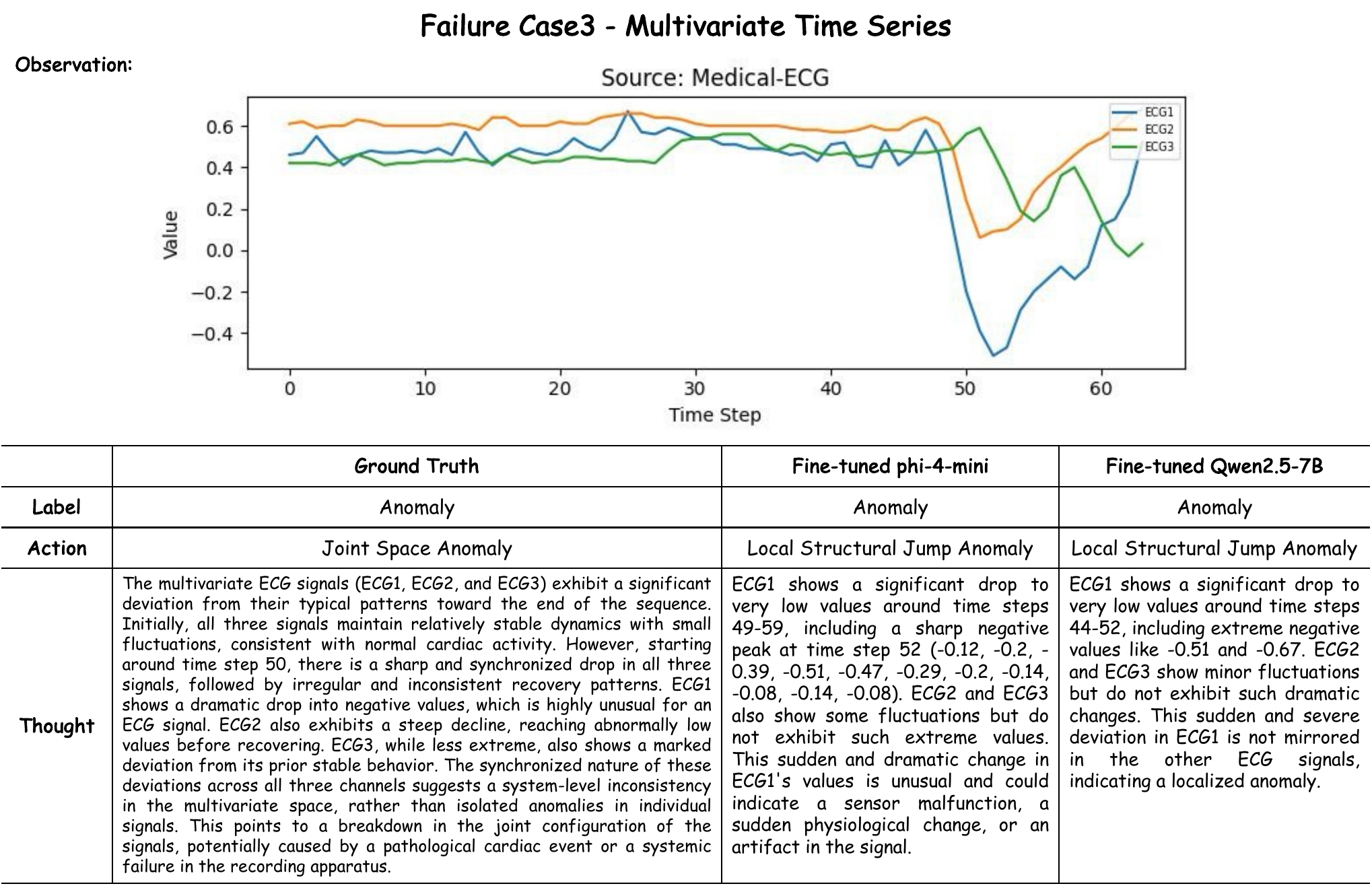}
    \caption{Multivariate time series failure case.}
    \label{fig:failure_example3}
\end{figure*}

\section{Failure Case Analysis} \label{Failure—case}
We further analyze representative failure cases to better understand the limitations of the proposed approach. There are three typical failure cases as shown in Figures~\ref{fig:failure_example1} - \ref{fig:failure_example3}. These failure cases provide additional insights into the boundaries of our approach rather than fundamental weaknesses. In failure case 1 (Figure~\ref{fig:failure_example1}), the anomaly manifests as a slow and smooth downward drift, which can be difficult to distinguish from normal long-term trends without explicit prior knowledge, leading to occasional under-detection. Failure case 2 (Figure~\ref{fig:failure_example2}) shows that while the model successfully detects anomalous behavior, it may confuse closely related anomaly types (e.g., periodic change versus amplitude variation) when multiple irregular patterns co-occur. Failure case 3 (Figure~\ref{fig:failure_example3}) highlights the challenge of multivariate reasoning under extreme local deviations, where a dominant abnormal channel can obscure or bias the interpretation of cross-variable relationships. Overall, these cases suggest that the model remains effective at identifying anomalous events, but finer-grained anomaly categorization and joint dependency modeling leave room for further improvement.

\section{Does Reinforcement Learning Improve Time Series Reasoning?}
We further investigate whether reinforcement learning (RL) can improve time series reasoning performance compared with supervised fine-tuning. We conduct a comparative study between SFT and GRPO-based RL on the \modelname ~task using Qwen2.5-7B. We perform experiments on the univariate dataset. The RL training uses a learning rate of $1\times10^{-6}$ and $\beta = 0.04$. 

We design two reward functions. \textbf{(1) Action Accuracy}: 1.0 for a correct match and 0.0 otherwise. \textbf{(2) Format Compliance}: 1.0 for including both \texttt{\textbackslash boxed1\{\}} and \texttt{\textbackslash boxed2\{\}}, 0.5 for including one, and 0.0 for none.

Table~\ref{tab:sft_grpo} shows the comparison between SFT and GRPO. While GRPO slightly improves label precision (0.8452 $\rightarrow$ 0.8571), it leads to consistent degradation in most metrics, especially Action F1 (0.1100 $\rightarrow$ 0.0872). This suggests that directly applying zero-initialized RL makes it difficult for the policy to explore and stabilize the complex reasoning paths required for fine-grained temporal categorization.

These results indicate that effective RL for time-series reasoning likely requires either a strong SFT warm-start or more granular process-level rewards to guide the model through intricate temporal logic.

\begin{table}[t]
\centering
\small
\caption{Performance comparison between SFT and GRPO on Qwen2.5-7B (Univariate dataset).}
\label{tab:sft_grpo}
\resizebox{\columnwidth}{!}{
\begin{tabular}{lcccc}
\toprule
\textbf{Metric} & \textbf{SFT} & \textbf{GRPO} & $\Delta$ \textbf{(GRPO-SFT)} & \textbf{Trend} \\
\midrule
Label Precision & 0.8452 & \textbf{0.8571} & +0.0119 & $\uparrow$ \\
Label Recall    & \textbf{0.9295} & 0.8872 & -0.0423 & $\downarrow$ \\
Label F1        & \textbf{0.8854} & 0.8719 & -0.0135 & $\downarrow$ \\
Action Precision& \textbf{0.2532} & 0.1256 & -0.1276 & $\downarrow$ \\
Action Recall   & 0.1229 & \textbf{0.1913} & +0.0684 & $\uparrow$ \\
Action F1       & \textbf{0.1100} & 0.0872 & -0.0228 & $\downarrow$ \\
\bottomrule
\end{tabular}}
\end{table}

\end{document}